\documentclass[11pt,a4paper]{article}

\usepackage{geometry}
\usepackage{coling2020}
\usepackage{times}
\usepackage{url}
\usepackage{latexsym}
\usepackage{microtype}
\usepackage[utf8x]{inputenc}
\usepackage{adjustbox}
\usepackage{booktabs}
\usepackage{multirow}
\usepackage{multicol}
\usepackage{comment}
\usepackage{eurosym}
\usepackage{covington}
\usepackage{enumitem}
\usepackage{subcaption}
\usepackage{caption}
\setlist[description]{noitemsep}
\usepackage{url}
\usepackage{csquotes}
\usepackage{amsthm,amsmath}
\usepackage{paralist}
\usepackage{xcolor}

\colingfinalcopy

\title{SemEval-2020 Task 1:\\Unsupervised Lexical Semantic Change Detection}

\author{Dominik Schlechtweg,$^{\clubsuit}$ Barbara McGillivray,$^{\diamondsuit,\heartsuit}$ Simon Hengchen,$^{\spadesuit}$\thanks{~~SH was affiliated with the University of Helsinki and the University of Geneva for most of this work.}\\ \textbf{Haim Dubossarsky,$^{\heartsuit}$ Nina Tahmasebi$^{\spadesuit}$} \\
{{\tt semeval2020lexicalsemanticchange@turing.ac.uk}} \\
$^{\clubsuit}$University of Stuttgart, $^{\diamondsuit}$The Alan Turing Institute, $^{\heartsuit}$University of Cambridge \\ $^{\spadesuit}$University of Gothenburg
}

\begin{document}

\maketitle

\begin{abstract}
Lexical Semantic Change detection, i.e., the task of identifying words that change meaning over time, is a very active research area, with applications in NLP, lexicography, and linguistics. Evaluation is currently the most pressing problem in Lexical Semantic Change detection, as no gold standards are available to the community, which hinders progress. We present the results of the first shared task that addresses this gap by providing researchers with an evaluation framework and manually annotated, high-quality datasets for English, German, Latin, and Swedish. 33 teams submitted 186 systems, which were evaluated on two subtasks.
\end{abstract}

\section{Overview}

Recent years have seen an exponentially rising interest in computational Lexical Semantic Change (LSC) detection \cite{ArXiV-Tahmasebi18,kutuzov-etal-2018-diachronic}. However, the field is lacking standard evaluation tasks and data. Almost all papers differ in how the evaluation is performed and what factors are considered in the evaluation. Very few are evaluated on a manually annotated diachronic corpus \cite[e.g.]{mcgillivray_2019,perrone_2019,Schlechtwegetal19}. This puts a damper on the development of computational models for LSC, and is a barrier for high-quality, comparable results that can be used in follow-up tasks. 

We report the results of the first SemEval shared task on Unsupervised LSC detection.\footnote{\url{https://languagechange.org/semeval}. In the remainder of this paper, \enquote{CodaLab} refers to this URL.} We introduce two related subtasks for computational LSC detection, which aim to identify the change in meaning of words over time using corpus data. We provide a high-quality multilingual (English, German, Latin, Swedish) LSC gold standard relying on approximately 100,000 instances of human judgment. For the first time, it is possible to compare the variety of proposed models on relatively solid grounds and across languages, and to put previously reached conclusions on trial. We may now provide answers to questions concerning the performance of different types of semantic representations (such as token embeddings vs. type embeddings, and topic models vs. vector space models), alignment methods and change measures. We provide a thorough analysis of the submitted results uncovering trends for models and opening perspectives for further improvements. In addition to this, the CodaLab website will remain open to allow any reader to directly and easily compare their results to the participating systems. We expect the long-term impact of the task to be significant, and hope to encourage the study of LSC in more languages than are currently studied, in particular less-resourced languages. 

\blfootnote{
    \hspace{-0.65cm}
    This work is licensed under a Creative Commons Attribution 4.0 International License. License details: \url{http://creativecommons.org/licenses/by/4.0/}.
}

\section{Subtasks}
\label{sec:task}

For the proposed tasks we rely on the comparison of two time-specific corpora $C_1$ and $C_2$. While this simplifies the LSC detection problem, it has two main advantages: (i) it reduces the number of time periods for which data has to be annotated, so we can annotate larger corpus samples and hence more reliably represent the sense distributions of target words; (ii) it reduces the task complexity, allowing different model architectures to be applied to it, widening the range of possible participants. 
Participants were asked to solve two subtasks:
\begin{description}
  \item[~~Subtask 1] Binary classification: for a set of target words, decide which words lost or gained sense(s) between $C_1$ and $C_2$, and which ones did not.
  \item[~~Subtask 2] Ranking: rank a set of target words according to their degree of LSC between $C_1$ and $C_2$.
\end{description}
For Subtask 1, consider the example of \emph{cell} in Table \ref{tab:SemChangeExample}, where the sense `phone' is newly acquired from $C_1$ to $C_2$ because its frequency is 0 in $C_1$ and $>0$ in $C_2$.
Subtask 2, instead, captures fine-grained changes in the two sense frequency distributions. For example, Table \ref{tab:SemChangeExample} shows that the frequency of the sense `chamber' drops from $C_1$ to $C_2$, although it is not totally lost. Such a change will increase the degree of LSC for Subtask 2, but will not count as change in Subtask 1. The notion of LSC underlying Subtask 1 is most relevant to historical linguistics and lexicography, while the majority of LSC detection models are rather designed to solve Subtask 2. Hence, we expected Subtask 1 to be a challenge for most models. Knowing whether, and to what degree a word has changed is crucial in other tasks, e.g. aiding in understanding historical documents, searching for relevant content, or historical sentiment analysis. The full LSC problem can be seen as a generalization of these two tasks into multiple time points where also the type of change needs to be identified.

\begin{table}[]
	\center
\small
	\begin{tabular}{ c | c c c | c  c c }
	\multirow{2}{*}{} &\multicolumn{3}{c |}{\textbf{$C_1$}} & \multicolumn{3}{c}{\textbf{$C_2$}}  \\	
	 \textbf{Senses} & \textbf{chamber} & \textbf{biology} & \textbf{phone} & \textbf{chamber} & \textbf{biology} & \textbf{phone} \\
	\# uses & 12 & 18 & 0 & 4 & 11 & 18 \\
	\end{tabular}
\caption{An example of a sense frequency distribution for the word \textit{cell} in $C_1$ and $C_2$.}\label{tab:SemChangeExample}
\vspace{-8ex}
\end{table}

\section{Data}

The task took place in a realistic unsupervised learning scenario. Participants were provided with trial and test data, but no training data. The public trial and test data consisted of a diachronic corpus pair and a set of target words for each language. Participants' predictions were evaluated against a set of hidden gold labels. The trial data consisted of small samples from the test corpora (see below) and four target words per language to which we assigned binary and graded gold labels randomly. Participants could not use this data to develop their models, but only to test the data input format and the online submission format. For development data participants were referred to three pre-existing diachronic data sets: DURel \cite{Schlechtwegetal18}, SemCor LSC \cite{SchlechtwegWalde20} and WSC \cite{Tahmasebi17}. In the evaluation phase participants were provided with the test corpora and a set of target words for each language.\footnote{\url{https://www.ims.uni-stuttgart.de/data/sem-eval-ulscd}}
Participants were asked to train their models only on the corpora described in Table \ref{tab:corpora}, though the use of pre-trained embeddings was allowed as long as they were trained in a completely unsupervised way, i.e., not on manually annotated data.

\subsection{Corpora} 
For English, we used the Clean Corpus of Historical American English (CCOHA) \cite{davies2002corpus,Alatrashetal20}, which spans 1810s--2000s. For German, we used the DTA corpus \cite{dta2017} and a combination of the BZ and ND corpora \cite{BZ2018,ND2018}. DTA contains texts from different genres spanning the 16th--20th centuries. BZ and ND are newspaper corpora jointly spanning 1945--1993. For Latin, we used the LatinISE corpus \cite{mcgillivray-kilgarriff} spanning from the 2nd century B.C. to the 21st century A.D. For Swedish, we used the Kubhist corpus \cite{KubHist}, a newspaper corpus containing texts from 18th--20th century.
The corpora are lemmatised and POS-tagged. CCOHA and DTA are spelling-normalized. BZ, ND and Kubhist contain frequent OCR errors \cite{adesam2019exploring,hengchen2020vocab}.

From each corpus we extracted two time-specific subcorpora $C_1$, $C_2$, as defined in Table \ref{tab:corpora}. The division was driven by considerations of data size and availability of target words (see below). From these two subcorpora we then sampled the released test corpora in the following way: Sentences with $<10$ tokens ($<2$ for Latin) were removed. German $C_2$ was downsampled to fit the size of $C_1$ by sampling all sentences containing target lemmas and combining them with a random sample of sentences not containing target lemmas of suited size. An equal procedure was applied to downsample English $C_1$ and $C_2$. For Latin and Swedish the full amount of sentences was used. Finally, all tokens were replaced by their lemma, punctuation was removed and sentences were randomly shuffled within each of $C_1$, $C_2$.\footnote{Sentence shuffling and lemmatization were done for copyright reasons. Participants were provided with start and end positions of sentences. Where Kubhist did not provide lemmatization (through KORP \cite{borin-etal-2012-korp}) we left tokens unlemmatized. Additional pre-processing steps were needed for English: for copyright reasons CCOHA contains frequent replacement tokens (10 x `@'). We split sentences around replacement tokens and removed them as a first step in the pre-processing pipeline. Further, because English frequently combines various POS in one lemma and many of our target words underwent POS-specific semantic changes, we concatenated targets in the English corpus with their broad POS tag (`target\_pos'). Also, the joint size of the CCOHA subcorpora had to be limited to $\sim$10M tokens because of copyright issues.} 
Find a summary of the released test corpora in Table \ref{tab:corpora}.

\begin{table}[]
\centering
\small
\begin{tabular}{|l|lllll|lllll|}
\hline
 &\multicolumn{5}{c|}{\textbf{$C_1$}} & \multicolumn{5}{c|}{\textbf{$C_2$}}  \\	
            & \textbf{corpus} & \textbf{period} & \textbf{tokens} & \textbf{types} & \textbf{TTR} & \textbf{corpus} & \textbf{period} & \textbf{tokens} & \textbf{types} & \textbf{TTR} \\
\hline
\textbf{English}  & CCOHA  & 1810--1860  & 6.5M  & 87k & 13.38 &  CCOHA & 1960--2010 & 6.7M & 150k & 22.38 \\
\textbf{German} & DTA & 1800--1899 & 70.2M & 1.0M & 14.25 & BZ+ND  & 1946--1990  & 72.3M & 2.3M  & 31.81\\
\textbf{Latin} & LatinISE & -200--0 & 1.7M & 65k & 38.24 & LatinISE & 0--2000& 9.4M  & 253k &  26.91 \\
\textbf{Swedish}   & Kubhist & 1790--1830 & 71.0M & 1.9M  & 47.88 & Kubhist & 1895--1903  & 110.0M & 3.4M & 17.27\\
\hline
\end{tabular}
\caption{Statistics of test corpora. TTR = Type-Token ratio (number of types / number of tokens * 1000)}\label{tab:corpora}
\vspace{-8ex} 
\end{table}

\subsection{Target words} 
Target words are either: (i) words that \emph{changed} their meaning(s) (lost or gained a sense) between $C_1$ and $C_2$; or (ii) \emph{stable} words that did not change their meaning during that time.\footnote{A target word is represented by its lemma form.} A large list of 100--200 changing words was selected by scanning etymological and historical dictionaries \cite{Paul02XXI,svenska,oxford2009oxford} for changes within the time periods of the respective corpora. This list was then further reduced by one annotator who checked whether there were meaning differences in samples of 50 uses from $C_1$ and $C_2$ per target word. Stable words were then chosen by sampling a control counterpart for each of the changing words with the same POS and comparable frequency development between $C_1$ and $C_2$, and manually verifying their diachronic stability as described above. Both types of words were annotated to obtain their sense frequency distributions as described below, which allowed us to verify the a-priori choice of changing and stable words. By balancing the target words for POS and frequency we aim to minimize the possibility that model biases towards these factors lead to artificially high performance \cite{dubossarsky2017,SchlechtwegWalde20}.

\subsection{Hidden/True Labels} 
For Subtask 1 (binary classification) each target word was assigned a binary label ($l\in \{0,1\}$) via manual annotation ($0$ for stable, $1$ for change). For Subtask 2 each target word was assigned a graded label ($0\leq l\leq 1$) according to their degree of LSC derived from the annotation ($0$ means no change, $1$ means total change). The hidden labels were published in the post-evaluation phase.\footnote{\url{https://www.ims.uni-stuttgart.de/data/sem-eval-ulscd-post}} Both types of labels (binary and graded) were derived from the sense frequency distributions of target words in $C_1$ and $C_2$ as obtained from the annotation process. For this, we adopt change notions similar to \newcite{SchlechtwegWalde20} as described below.

\section{Annotation}
\label{sec:annotation}

We focused our efforts on annotating large and more representative samples for a limited number of words rather than annotating many words.\footnote{An indication that random samples with the chosen sizes can indeed be expected to be representative of the population is given by the results of the simulation study described in Appendix \ref{sec:example}: We were able to nearly fully recover the population clustering structure from the samples (average of $>.96$ adjusted mean rand index).} In this section we describe the setup of the annotation for the modern languages (English, German, and Swedish) first. The setup for Latin is slightly different and we describe it later in this section. 

We started with four annotators per language, but had to add additional annotators later because of a high annotation load and dropouts. The total number of annotators for English/German/Swedish was 9/8/5. All annotators were native speakers and present or former university students. For German we had two annotators with a background in historical linguistics, while for English and Swedish we had one such annotator. For each target word we randomly sampled 100 uses from each of $C_1$ and $C_2$ for annotation (total of 200 uses per target word).\footnote{We refer to an occurrence of a word $w$ in a sentence by `use of $w$'.} If a target word had less than 100 uses, we annotated the full sample. We then mixed the use samples of a target word into a joint set $U$ and annotated $U$ using an extension of the DURel framework \cite{Schlechtwegetal18,haettySurel:2019,Erk13}. DURel produces high inter-annotator agreement even between non-expert annotators relying on the simple notion of semantic relatedness. Pairs of word uses from $C_1$ and $C_2$ are annotated on a four-point scale from unrelated meanings (1) to identical meanings (4) (see Table \ref{tab:scales}). Our extension consisted in the sampling procedure of use pairs: instead of annotating a random sample of pairs and using comparison of their mean relatedness over time as a measure of LSC \cite{Schlechtwegetal18}, we aimed to sample pairs such that after annotation they span a sparsely connected \textbf{usage graph} combining the uses from $C_1$, $C_2$, where nodes represent uses and edges represent (the median of) annotator judgments (see Figure \ref{fig:graph1}). This usage graph was then clustered into sets of uses expressing the same sense \cite{Schutze1998}. By further distinguishing two subgraphs for $C_1$, $C_2$ we got two clusterings with a shared set of clusters, because they were obtained on the same total graph \cite{palla2007quantifying}. We then equated the two clusterings obtained for $C_1$, $C_2$ with their respective sense frequency distributions $D_1$, $D_2$. The change scores followed immediately (see below). Note that this extension remained hidden from the annotators: as with DURel their only task was to judge the relatedness of use pairs. These were presented to annotators in randomized order.

\subsection{Edge sampling} 
\label{sec:edge}
Retrieving the full usage graph is not feasible even for a small set of $n$ uses as this implies annotating $n*(n-1)/2$ edges. Hence, the main challenge with our annotation approach was to reduce the number of edges to annotate as much as possible, while keeping the necessary information needed to infer a meaningful clustering on the graph. We did this by annotating the data in several rounds. After each round the usage graph of a target word was updated with the new annotations and a new clustering was obtained.\footnote{If an edge was annotated by several annotators we took the median as an edge weight.} Based on this clustering we sampled the edges for the next round applying simple heuristics similar to \newcite{Biemann-2013-system}, a detailed description of which can be found in Appendix \ref{sec:example}. We spread the annotation load randomly over annotators making sure that roughly half of the use pairs were annotated by more than one annotator. 

\begin{table}[]
\parbox{.45\linewidth}{
\centering
\tabcolsep=0.11cm
\begin{tabular}{ll}
\multirow{4}{*}{$\Bigg\uparrow$}&Identity\\
&Context Variance\\
&Polysemy\\
&Homonymy
\end{tabular}
\label{tab:blank}}
\hfill
\parbox{.45\linewidth}{
\centering
\begin{tabular}{ll}
\multirow{4}{*}{$\Bigg\uparrow$} &4: Identical\\
 &3: Closely Related\\
 &2: Distantly Related\\
 &1: Unrelated\\
\end{tabular}
\label{tab:scale2}}
\caption{\newcite{Blank97XVI}'s continuum of semantic proximity (left) and the DURel relatedness scale derived from it (right).}\label{tab:scales}
\vspace{-8ex} 
\end{table}

\subsection{Special Treatment of Latin} 
Latin poses a special case due to the lack of native speakers. We recruited 10 annotators with a high-level knowledge of Latin, and ranging from undergraduate students to PhD students, post-doctoral researchers, and more senior researchers. We selected a range of target words whose meaning had changed between the pre-Christian and the Christian era according to the literature \cite{clackson} and in the pre-annotation trial we checked that both meanings were present in the corpus data. For each changed word, we selected a control word whose meaning did not change from the pre-Christian era and the Christian era, whose PoS was the same as the changed word, and whose frequency values in each of the two subcorpora ($f_{c_{c_1}}$ and $f_{c_{c_2}}$) were in the following intervals: $f_{c_{c_1}}\in[f_{t_{c_1}}-p*f_{c_{t_1}},f_{t_{c_1}}+p*f_{c_{t_1}}]$ and $f_{c_{c_2}}\in[f_{t_{c_2}}-p*f_{c_{t_2}},f_{t_{c_2}}+p*f_{c_{t_2}}]$, respectively, where $p$ ranged between 0.03 and 0.15 and $f_{t_{c_1}}$ and $f_{t_{c_2}}$ are the frequency of the changed word in $C_1$, $C_2$.\footnote{We experimented with increasing values of $p$ and chose the minimum for which a control word could be found for each changed word.}
In a trial annotation task our annotators reported difficulties and that they had to translate to their native language when comparing two excerpts of text. Hence, we decided to use a variation of the procedure described above which was introduced by \newcite{Erk13}. Instead of use pairs, annotators judged the relatedness between a use and a sense definition from a dictionary, on the DURel scale. The sense definitions were taken from the Latin portion of the Logeion online dictionary.\footnote{\url{https://logeion.uchicago.edu/}.} We selected 30 sample sentences for each of $C_1$, $C_2$. Due to the challenge of finding qualified annotators, each word was assigned only to one annotator. We treated sense definitions as additional nodes in a usage graph connected to uses by edges representing annotator judgments. Clustering was then performed as for the other languages.

\begin{table}[]
\centering
\small
\begin{tabular}{|l|lllll|llll|llll|}
\hline
& \multicolumn{5}{c |}{\textbf{General}} &\multicolumn{4}{c |}{\textbf{Subtask 1}} & \multicolumn{4}{c|}{\textbf{Subtask 2}} \\
        & $n$ & N/V/A & AGR & LOSS & JUD & LSC & FRQ\textsubscript{d} & FRQ\textsubscript{m} & PLY\textsubscript{m} & LSC & FRQ\textsubscript{d} & FRQ\textsubscript{m} & PLY\textsubscript{m} \\
\hline
\textbf{English}  & 37 & 33/4/0 & .69 & .27 & 30k & .43 & -.18 & -.03 & .45 & .24 & -.29 & -.05  & .72\\
\textbf{German}  & 48 & 32/14/2 & .59 & .20 & 38k & .35 & -.06 & -.11 &  .68 & .31 & .00 & -.02 & .73  \\
\textbf{Latin}  & 40 & 27/5/8 & - & .26 & 9k & .65 & .16 & .02 & .14 & .33 & .39 & -.13  & .31   \\
\textbf{Swedish}  &  31 &  23/5/3 & .58 & .12 & 20k & .26 & -.04 & -.29 & .45 & .16 & .00 & -.13  & .75  \\
\hline
\end{tabular}
\caption{Overview target words. $n$ = number of target words, N/V/A = number of nouns/verbs/adjectives, AGR = inter-annotator agreement in round 1, LOSS = mean of normalized clustering loss * 10, JUD = number of judged use pairs, LSC = mean binary/graded change score, FRQ\textsubscript{d} = Spearman correlation between change scores and target words' absolute difference in log-frequency between $C_1$, $C_2$. Similarly for minimum frequency (FRQ\textsubscript{m}) and minimum number of senses (PLY\textsubscript{m}) across $C_1$, $C_2$. }
\label{tab:targetstats}
\vspace{-8ex} 
\end{table}

\subsection{Clustering}
\label{sec:clustering}

The usage graphs we obtain from the annotation are weighted, undirected, sparsely observed and noisy. This poses a very specific problem that calls for a robust clustering algorithm. For this, we rely on a variation of correlation clustering \cite{Bansal04} by minimizing the sum of cluster disagreements, i.e., the sum of negative edge weights within a cluster plus the sum of positive edge weights across clusters. To see this, consider \newcite{Blank97XVI}'s continuum of semantic proximity and the DURel relatedness scale derived from it, as illustrated in Table \ref{tab:scales}. In line with Blank, we assume that use pairs with judgments of 3 and 4 are more likely to belong to the same sense, while judgments of 1 and 2 are more likely to belong to different senses. Consequently, we shift the weight $W(e)$ of all edges $e \in E$ in a usage graph $\mathbf{G = (U, E, W)}$ by $W(e)-2.5$. We refer to those edges $e \in E$ with a weight $W(e) \geq 0$ as \textbf{positive} edges $P_E$ and edges with weights $W(e)<0$ as \textbf{negative} edges $N_E$. Let further $C$ be some clustering on $U$, $\phi_{E,C}$ be the set of positive edges \textbf{across} any of the clusters in clustering $C$ and $\psi_{E,C}$ the set of negative edges \textbf{within} any of the clusters. We then search for a clustering $C$ that minimizes $L(C)$:
\begin{equation}\label{eq:loss}
L(C) = \sum_{e\in \phi_{E,C}} W(e) + \sum_{e\in \psi_{E,C}} |W(e)| ~~.
\end{equation}
That is, we try to minimize the sum of positive edge weights between clusters and (absolute) negative edge weights within clusters. Minimizing $L$ is a discrete optimization problem which is NP-hard \cite{Bansal04}. However, we have a relatively low number of nodes ($\leq 200$), and hence, the global optimum can be approximated sufficiently with a standard optimization algorithm. We choose Simulated Annealing \cite{Pincus1970} as we do not have strong efficiency constraints and the algorithm showed superior performance in a simulation study. More details on the procedure can be found in Appendix~\ref{sec:example}. In order to reduce the search space, we iterate over different values for the maximum number of clusters. We also iterate over randomly as well as heuristically chosen initial clustering states.\footnote{We used mlrose to perform the clustering \cite{Hayes19}.}

This way of clustering usage graphs has several advantages: (i) It finds the optimal number of clusters on its own. (ii) It easily handles missing information (non-observed edges). (iii) It is robust to errors by using the global information on the graph. That is, a wrong judgment can be outweighed by correct ones. (iv) It directly optimizes an intuitive quality criterion on usage graphs. Many other clustering algorithms such as Chinese Whispers \cite{Biemann2006} make local decisions, so that the final solution is not guaranteed to optimize a global criterion such as $L$. (v) By weighing each edge with its (shifted) weight, $L$ respects the gradedness of word meaning. That is, edges with $|W(e)| \approx 0$ have less influence on $L$ than edges with $|W(e)| \approx 1.5$. Finally, it showed superior performance to all other clustering algorithms we tested in a simulation study. (See Appendix \ref{sec:example}.)
\begin{figure}[]
    \begin{subfigure}{0.33\textwidth}
\frame{        \includegraphics[width=\linewidth]{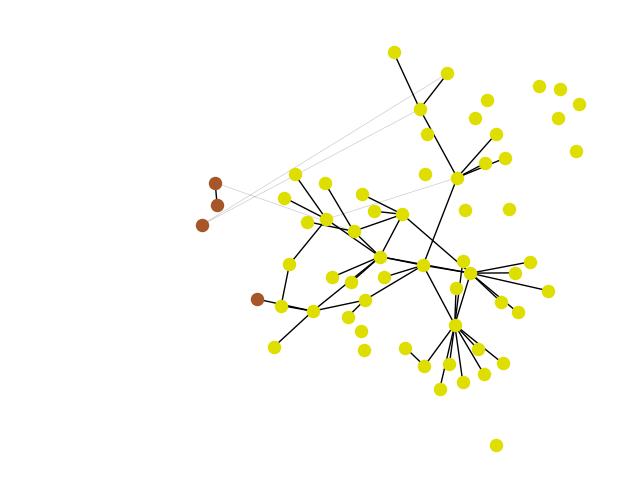}}
        \caption*{$C_1$}
    \end{subfigure}
    \begin{subfigure}{0.33\textwidth}
\frame{        \includegraphics[width=\linewidth]{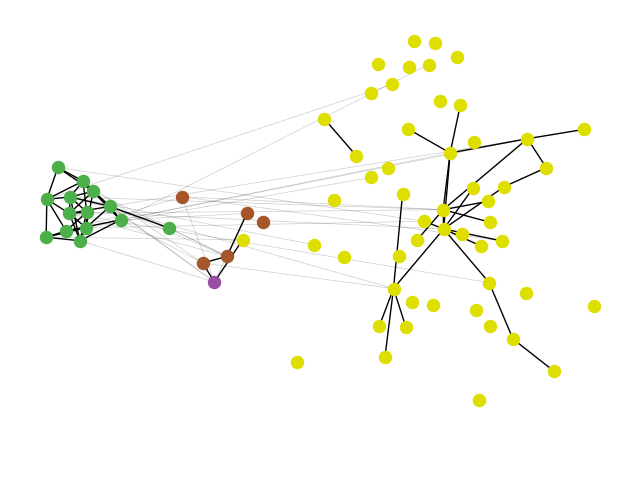}}
        \caption*{$C_2$}
    \end{subfigure}
    \begin{subfigure}{0.33\textwidth}
\frame {        \includegraphics[width=\linewidth]{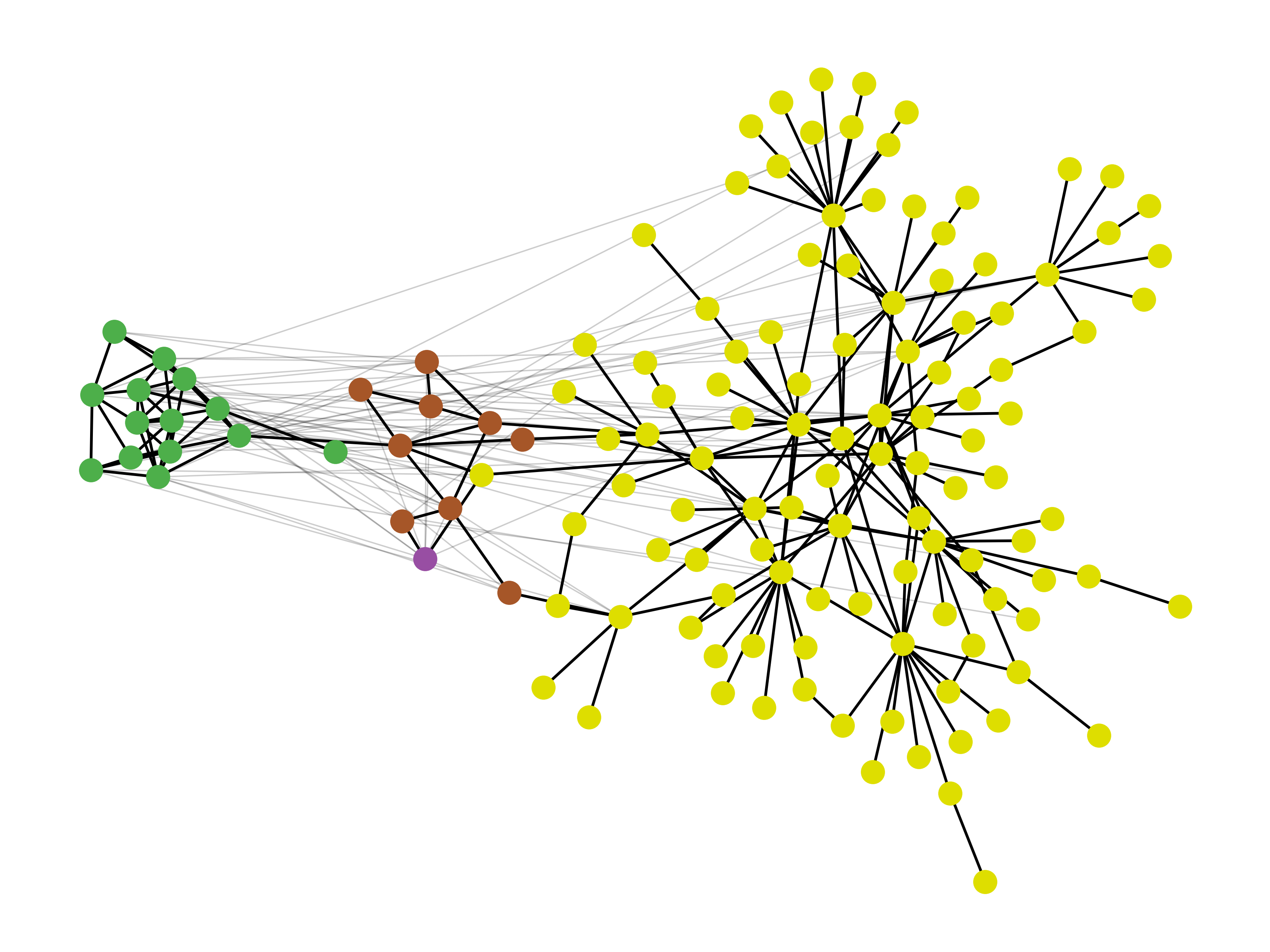}}
        \caption*{full}
    \end{subfigure}
    \caption{Usage graph of Swedish \textit{ledning}.  $D_1=(58,0,4,0)$, $D_2=(52,14,5,1)$, $B(w)=1$ and $G(w)=0.34$.}\label{fig:graph1}
\vspace{-7ex}
\end{figure}

\subsection{Change scores} 
\label{sec:scores}
A sense frequency distribution (SFD) encodes how often a word $w$ occurs in each of its senses \cite[e.g.]{mccarthy2004,lau-EtAl:2014}. From the clustering we obtain two SFDs $D$, $E$ for a word $w$ in the two corpora $C_1$, $C_2$, where each cluster corresponds to one sense.\footnote{The frequency for sense $i$ in corpus $C$ is given by the number of uses from $C$ in the cluster corresponding to sense $i$.} Binary LSC for Subtask 1 of the word $w$ is then defined as
\begin{equation}\label{eq:binary}
  \begin{split}
  B(w)= 1 & \textnormal{ if for some $i$, ${D}_i \leq k$ and ${E}_i \geq n$,}\\
    & \textnormal{or vice versa.}\\
  B(w)= 0 & \textnormal{ else.}
  \end{split}
\end{equation}
where $D_i$ and $E_i$ are the frequencies of sense $i$ in $C_1$, $C_2$ and $k$, $n$ are lower frequency thresholds aimed to avoid that small random fluctuations in sense frequencies caused by sampling variability or annotation error are misclassified as change \cite{SchlechtwegWalde20}. According to Definition \ref{eq:binary}, a word is classified as gaining a sense, if the sense is attested at most $k$ times in the annotation sample from $C_1$, but attested at least $n$ times in the sample from $C_2$. (Similarly for words that lose a sense.) We set $k=0$, $n=1$ for the smaller samples ($\leq 30$) in Latin and $k=2$, $n=5$ for the larger samples ($\leq 100$) in English, German, Swedish. We make no distinction between words that gain vs. words that lose senses, both fall into the change class. Equally, we make no distinction between words that gain/lose one sense vs. words that gain/lose several senses.

For graded LSC in Subtask 2 we first normalize $D$ and $E$ to probability distributions $P$ and $Q$ by dividing each element by the total sum of the frequencies of all senses in the respective distribution. The degree of LSC of the word $w$ is then defined as the Jensen-Shannon distance between the two normalized frequency distributions:
\begin{equation}
G(w)=JSD(P,Q)
\end{equation}
where the Jensen-Shannon distance is the symmetrized square root of the Kullback-Leibler divergence \cite{Lin1991,DonosoS17}. $G(w)$ is symmetric, ranges between $0$ and $1$ and is high if $P$ and $Q$ assign very different probabilities to the same senses. Note that $B(w)$ and $G(w)$ not necessarily correspond to each other: a word $w$ may show no binary change but high graded change, or vice versa.

\begin{figure}[]
    \begin{subfigure}{0.33\textwidth}
\frame{        \includegraphics[width=\linewidth]{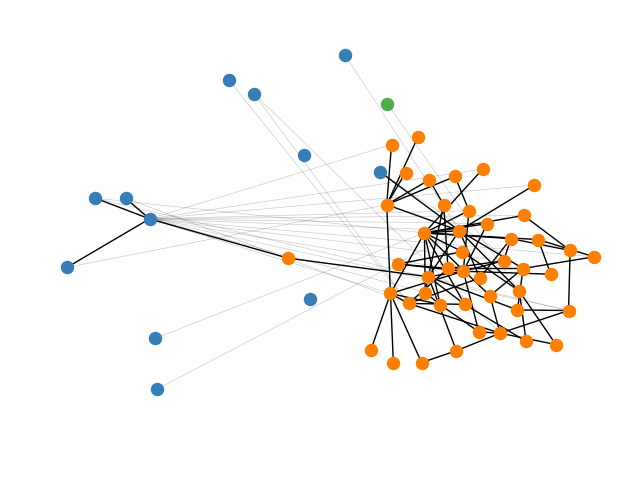}}
        \caption*{$C_1$}
    \end{subfigure}
    \begin{subfigure}{0.33\textwidth}
\frame{        \includegraphics[width=\linewidth]{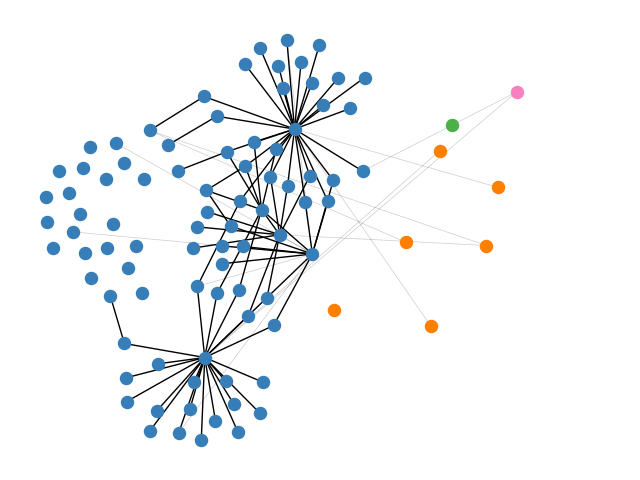}}
        \caption*{$C_2$}
    \end{subfigure}
    \begin{subfigure}{0.33\textwidth}
\frame {        \includegraphics[width=\linewidth]{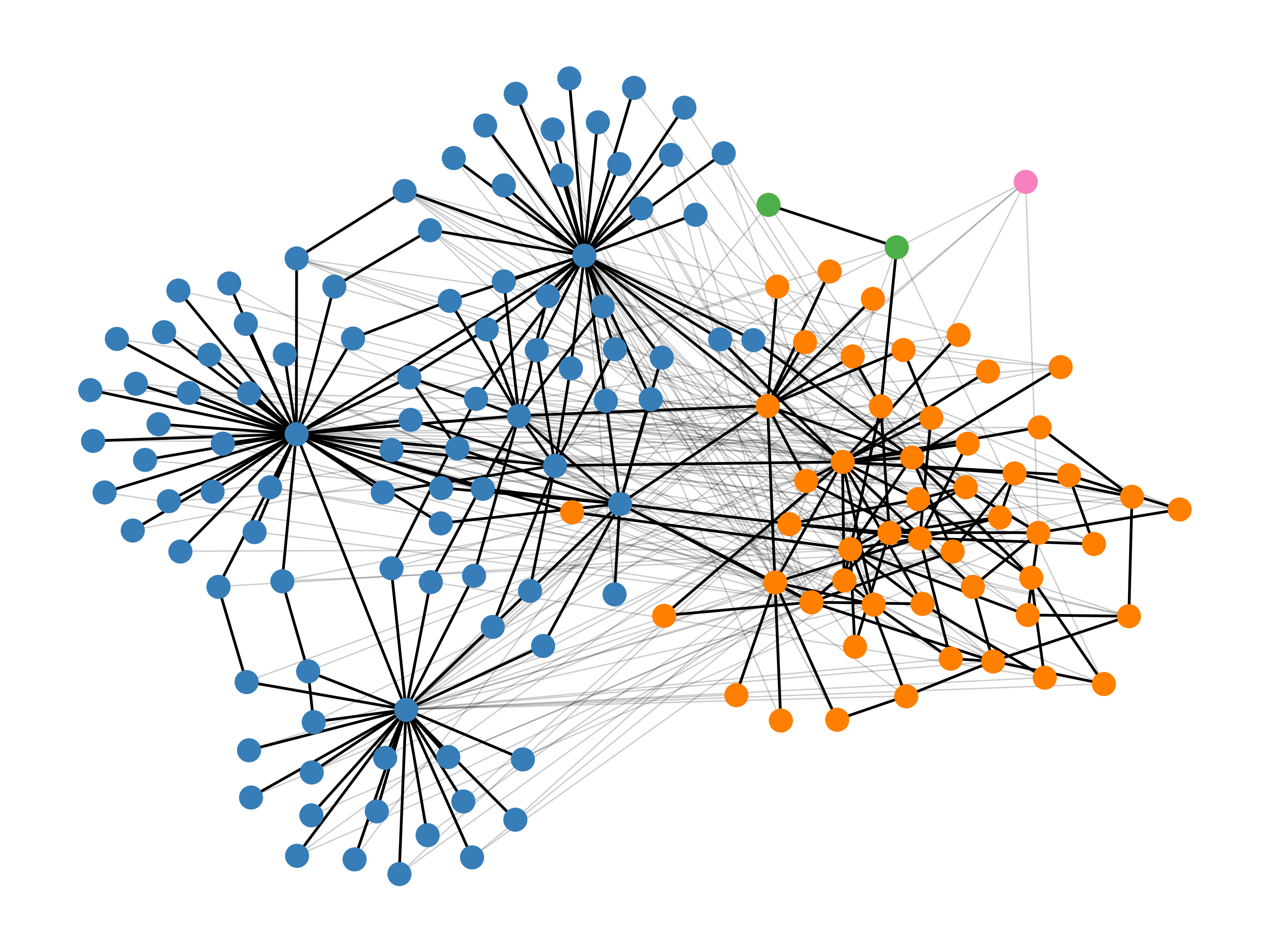}}
        \caption*{full}
    \end{subfigure}
    \caption{Usage graph of German \textit{Eintagsfliege}. $D_1=( 12, 45, 0, 1)$, $D_2=( 85, 6, 1, 1)$, $B(w)=0$ and $G(w)=0.66$.}\label{fig:graph2}
\vspace{-7ex}
\end{figure}

\subsection{Result} 

Figure \ref{fig:graph1} and Figure \ref{fig:graph2} show the annotated and clustered usage graphs for Swedish target \textit{ledning} and German target \textit{Eintagsfliege}. Nodes represent uses of the target word.
Edges represent the median of relatedness judgments between uses (\textbf{black}/\textcolor{gray}{gray} lines for \textbf{positive}/\textcolor{gray}{negative} edges). Colors make clusters (senses) inferred on the full graph. After splitting the full graph into the two time-specific subgraphs for $C_1$, $C_2$ we obtain the two sense frequency distributions $D_1$, $D_2$. From these we inferred the binary and the graded change value. The two words represent semantic changes indicative of Subtask 1 and 2 respectively: \textit{ledning} gains a sense with rather low frequency in $C_2$. Hence, it has binary change, but low graded change. For \textit{Eintagsfliege}, however, its two main senses exist in both $C_1$ and $C_2$, while their frequencies change dramatically. Hence, it has no binary change, but high graded change.

Find a summary of the annotation outcome for all languages and target words in Table \ref{tab:targetstats}. The final test sets contain between 31 (Swedish) and 48 (German) target words. Throughout the annotation we excluded several targets if they had a high number of `0' judgments or needed a high number of further edges to be annotated. As previous studies, we report the mean of Spearman correlations between annotator judgments as agreement measure. \newcite{Erk13} and \newcite{Schlechtwegetal18} report agreement scores between 0.55 and 0.68, which is comparable to our scores.\footnote{Note that because we spread disagreements from previous rounds in each round to further annotators, on average uses in later rounds become much harder to judge, which has a negative effect on agreement. Hence, for comparability reasons we report the agreement in the first round where no disagreement detection has taken place. The agreement across all rounds, calculated as weighted mean of agreements is 0.52/0.60/-/0.58.} The clustering loss is the value of $L$ (Definition \ref{eq:loss}) divided by the maximum possible loss on the respective graph. It gives a measure of how well the graphs could be partitioned into clusters by the $L$ criterion.

The class distribution (column `LSC') for Subtask 1 differs per language as a result of several target words being dropped during the annotation. In Latin the majority of target words have binary change, while in Swedish the majority has no binary change. This is also reflected in the mean scores for graded LSC in Subtask 2. Despite the excluded target words the frequency statistics are roughly balanced (FRQ\textsubscript{d}, FRQ\textsubscript{m}). However, we did not control the test sets for polysemy and there are strong correlations for English, German and Swedish between graded change and polysemy in Subtask 2 (PLY\textsubscript{m}). This correlation reduces for binary change in Subtask 1 but is still moderate for English and Swedish and remains high for German.

In total, roughly 100,000 judgments were made by annotators. For English/German/Swedish $\approx $ 50\% of the use pairs were annotated by more than one annotator. In total, the annotation cost roughly \EUR{20,000} for 1,000 hours -- twice as much as originally budgeted.

\section{Evaluation}

All teams were allowed a total of 10 submissions, the best of which was kept for the final ranking in the competition. Participants had to submit predictions for both subtasks and all languages. A submission's final score for each subtask was computed as the average performance across all four languages. During the evaluation phase, the leaderboard was hidden, as per SemEval recommendation. 

\subsection{Scoring Measures}
For Subtask 1 submitted predictions were evaluated against the hidden labels via accuracy, given that we anticipated the class distribution for target words to be approximately balanced before the annotation. Scores are bounded between $0$ and $1$. As the distribution turned out to be imbalanced for some languages, we also report F1-score in Appendix \ref{A:results}. 
For Subtask 2, we used Spearman's rank-order correlation coefficient $\rho$ with the gold rank. Spearman's $\rho$ only considers the order of the words, the actual predicted change values were not taken into account. Ties are corrected by assigning the average of the ranks that would have been assigned to all the tied values to each value (e.g. two words sharing rank 1 both get assigned rank 1.5). Scores are bounded between $-1$ (completely opposite to true ranking) and $1$ (exact match). 
\subsection{Baselines} 

For both subtasks, we have two baselines: (i) Normalized frequency difference (Freq. Baseline) first calculates the frequency for each target word in each of the two corpora, normalizes it by the total corpus frequency and then calculates the absolute difference between these values as a measure of change. (ii) Count vectors with column intersection and cosine distance (Count Baseline) first learns vector representations for each of the two corpora, then aligns them by intersecting their columns and measures change by cosine distance between the two vectors for a target word. 
A Python implementation of both these baselines was provided in the starting kit.
A third baseline, for Subtask 1, is the majority class prediction (Maj. Baseline), i.e., always predicting the `0' class (no change). 

\section{Participating Systems}

Thirty-three teams participated in the task, totaling 53 members. The teams submitted a total of 186 submissions. Given the large number of teams, we provide a summary of the systems in the body of this paper.  A more detailed description of each participating system for which a paper was submitted is available in Appendix~\ref{A:systems}. We also encourage the reader to read the full system description papers.

Participating models can be described as a combination of (i) a semantic representation, (ii) an alignment technique and (iii) a change measure.
Semantic representations are mainly average embeddings (\emph{type} embeddings) and contextualized embeddings (\emph{token} embeddings). Token embeddings are often combined with a clustering algorithm such as K-means, affinity propagation \cite{frey2007clustering}, (H)DBSCAN, GMM, or agglomerative clustering.
One team uses a graph-based semantic network, one a topic model and several teams also propose ensemble models.
Alignment techniques include Orthogonal Procrustes \cite[OP]{Hamilton:2016}, Vector Initialization \cite[VI]{Kim14}, versions of Temporal Referencing \cite[TR]{Dubossarskyetal19}, and Canonical Correlation Analysis (CCA). 
A variety of change measures are applied, including Cosine Distance (CD), Euclidean Distance (ED), Local Neighborhood Distance (LND), Kullback-Leibler Divergence (KLD), mean/standard deviation of co-occurrence vectors, or cluster frequency.
Table~\ref{tab:results} shows the type of system for every team's best submission for both subtasks.

\section{Results}
\begin{table}

\small
\centering
\tabcolsep=0.09cm
\begin{tabular}{l|rrrrr|c}
\toprule
 \multirow{ 2}{*}{\textbf{Team}}&\multicolumn{5}{c|}{\textbf{Subtask 1}}& \multirow{ 2}{*}{\textbf{System}}\\
{} &  Avg. &    EN &    DE &    LA &    SV & \\
\midrule
UWB                &  .687 &  .622 &  .750 &  .700 &  .677 &type\\
Life-Language      &  .686 &  .703 &  .750 &  .550 &  .742&type \\
Jiaxin \& Jinan     &  .665 &  .649 &  .729 &  .700 &  .581 &type\\
RPI-Trust          &  .660 &  .649 &  .750 &  .500 &  .742 &type\\
UG\_Student\_Intern  &  .639 &  .568 &  .729 &  .550 &  .710 &type\\
DCC                &  .637 &  .649 &  .667 &  .525 &  .710&type \\
NLP@IDSIA          &  .637 &  .622 &  .625 &  .625 &  .677 &token\\
JCT                &  .636 &  .649 &  .688 &  .500 &  .710 &type\\
Skurt              &  .629 &  .568 &  .562 &  .675 &  .710 &token\\
Discovery\_Team     &  .621 &  .568 &  .688 &  .550 &  .677 &ens.\\
\textbf{Count Bas.} &  .613 &  .595 &  .688 &  .525 &  .645 & - \\
TUE                &  .612 &  .568 &  .583 &  .650 &  .645&token \\
Entity             &  .599 &  .676 &  .667 &  .475 &  .581 &type\\
IMS                &  .598 &  .541 &  .688 &  .550 &  .613 &type\\
cs2020             &  .587 &  .595 &  .500 &  .575 &  .677&token \\
UiO-UvA            &  .587 &  .541 &  .646 &  .450 &  .710 &token\\
NLPCR              &  .584 &  .730 &  .542 &  .450 &  .613 &token\\
\textbf{Maj. Bas.}  &  .576 &  .568 &  .646 &  .350 &  .742 & -\\
cbk                &  .554 &  .568 &  .625 &  .475 &  .548 &token\\
Random             &  .554 &  .486 &  .479 &  .475 &  .774 &type\\
UoB                &  .526 &  .568 &  .479 &  .575 &  .484 &topic\\
UCD                &  .521 &  .622 &  .500 &  .350 &  .613 &graph\\
RIJP               &  .511 &  .541 &  .500 &  .550 &  .452 &type\\
\textbf{Freq. Bas.} &  .439 &  .432 &  .417 &  .650 &  .258 & - \\
\bottomrule
\end{tabular}
\quad
\begin{tabular}{l|rrrrr|c}
\toprule
 \multirow{ 2}{*}{\textbf{Team}}&\multicolumn{5}{c|}{\textbf{Subtask 2}}& \multirow{ 2}{*}{\textbf{System}}\\
{} &  Avg. &    EN &    DE &    LA &    SV & \\
\midrule
UG\_Student\_Intern  &  .527 &  .422 &  .725 &  .412 &  .547 &type\\
Jiaxin \& Jinan     &  .518 &  .325 &  .717 &  .440 &  .588 &type\\
cs2020             &  .503 &  .375 &  .702 &  .399 &  .536 &type\\
UWB                &  .481 &  .367 &  .697 &  .254 &  .604 &type\\
Discovery\_Team     &  .442 &  .361 &  .603 &  .460 &  .343&ens. \\
RPI-Trust          &  .427 &  .228 &  .520 &  .462 &  .498 &type\\
Skurt              &  .374 &  .209 &  .656 &  .399 &  .234&token \\
IMS                &  .372 &  .301 &  .659 &  .098 &  .432&type \\
UiO-UvA            &  .370 &  .136 &  .695 &  .370 &  .278&token \\
Entity             &  .352 &  .250 &  .499 &  .303 &  .357 &type\\
Random             &  .296 &  .211 &  .337 &  .253 &  .385&type \\
NLPCR              &  .287 &  .436 &  .446 &  .151 &  .114 &token\\
JCT                &  .254 &  .014 &  .506 &  .419 &  .078&type \\
cbk                &  .234 &  .059 &  .400 &  .341 &  .136&token \\
UCD                &  .234 &  .307 &  .216 &  .069 &  .344 &graph\\
Life-Language      &  .218 &  .299 &  .208 & -.024 &  .391 &type\\
NLP@IDSIA          &  .194 &  .028 &  .176 &  .253 &  .321&token \\
\textbf{Count Bas.} &  .144 &  .022 &  .216 &  .359 & -.022 & -\\
UoB                &  .100 &  .105 &  .220 & -.024 &  .102 &topic\\
RIJP               &  .087 &  .157 &  .099 &  .065 &  .028&type \\
TUE                &  .087 & -.155 &  .388 &  .177 & -.062 &token\\
DCC                & -.083 & -.217 &  .014 &  .020 & -.150 &type\\
\textbf{Freq. Bas.} & -.083 & -.217 &  .014 &  .020 & -.150 & - \\
\textbf{Maj. Bas.}  &    - &    - &    - &    - &    -  & -\\
\bottomrule
\end{tabular}
\caption{Summary of the performance of systems for which a system description paper was submitted, as well as their type of semantic representation for that specific submission in Subtask 1 (left) and Subtask 2 (right). For each team, we report the values of accuracy (Subtask 1) and Spearman correlation (Subtask 2) corresponding to their best submission in the evaluation phase. Abbreviations: Avg. = average across languages, EN = English, DE = German, LA = Latin, and SV = Swedish, type = average embeddings, token = contextualised embeddings, topic = topic model, ens. = ensemble, graph = graph, UCD = University\_College\_Dublin.}\label{tab:results}
\vspace{-8ex} 
\end{table}

As illustrated by Table~\ref{tab:results}, \textbf{UWB} has the best performance in Subtask 1 for the average over all languages, closely followed by \textbf{Life-Language}, \textbf{Jiaxin \& Jinan}\footnote{The team is named \enquote{LYNX} on the competition CodaLab.} and \textbf{RPI-Trust}.\footnote{The team submits an ensemble model. As all of the features are derived from the type vectors, we classify it as ``type" in this section.}
For Subtask 2, \textbf{UG\_Student\_Intern} performs best, followed by \textbf{Jiaxin \& Jinan} and \textbf{cs2020}.\footnote{The team is named \enquote{cs2020} and \enquote{cs2021} on the competition CodaLab. The combined number of submissions made by the two teams did not exceed the limit of 10.} Across all systems, good performance in Subtask 1 does not indicate good performance in Subtask 2 (correlation between the system ranks is 0.22). However, and with the exception of \textbf{Life-Language} and \textbf{cs2020}, most top performing systems in Subtask 1 also excel in Subtask 2, albeit with a slight change of ranking. 

Remarkably, all the top performing systems use static-type embedding models, and differ only in terms of their solutions to the alignment problem (Canonical Correlation Analysis, Orthogonal Procrustes, or Temporal Referencing). Interestingly, the top systems refine their models using one or more of the following steps: a) computing additional features from the embedding space; b) combining scores from different models (or extracted features) using ensemble models; c) choosing a threshold for changed words based on a distribution of change scores. We conjecture that these additional (and sometimes very original) post-processing steps are crucial for these systems' success. We now briefly describe the top performing systems in terms of these three steps (for further details please see Appendix~\ref{A:systems}). \textbf{UWB} (SGNS+CCA+CD) sets the average change score as the threshold (c). \textbf{Life-Language} (SGNS) represents words according to their distances to a set of stable pivot words in two unaligned spaces, and compares their divergence relative to a distribution of change scores obtained from unstable pivot words (a+c). \textbf{RPI-Trust} (SGNS+OP) extract features (a word's cosine distance, change of distances to its nearest-neighbours and change in frequency), transform each word's feature to a CDF score, and averages these probabilities (a+b+c). \textbf{Jiaxin \& Jinan} (SGNS+TR+CD) fits the empirical cosine distance change scores to a Gamma Quantile Threshold, and sets the 75\% quantile as the threshold (c). \textbf{UG\_Student\_Intern} (SGNS+OP) measures change using Euclidean distance instead of cosine distance. \textbf{cs2020} uses SGNS+OP+CD only as baseline method. 

An important finding common to most systems is the difference between their performances across the four languages --  systems that excel in one language do not necessarily perform well in another. This discrepancy may be due to a range of factors, including the difference in corpus size and the nature of the corpus data, as well as the relative availability of resources in some languages such as English over others. The Latin corpus, for example, covers a very long time span, and the lower performance of the systems on this language may be explained by the fact that the techniques employed, especially word token/type embeddings, have been developed for living languages and little research is available on their adaptation to dead and ancient languages. In general, dead languages tend to pose additional challenges compared to living languages \cite{piotrowski}, due to a variety of factors, including their less-resourced status, lack of native speakers, high linguistic variation and non-standardized spelling, and errors in Optical Character Recognition (OCR). Other factors that should be investigated are data quality \cite{hill2019quantifying,van2020assessing}: while English and Latin are clean data, German and Swedish present notorious OCR errors.
The availability of tuned hyperparameters might have played a role as well: for German, some teams report following prior work such as \newcite{Schlechtwegetal19}.
Finally, another factor for the discrepancy in performance between languages for any given system is not related to the nature of the systems nor of the data, but due to the fact that some teams focused on some languages, submitting dummy results for the others.

\paragraph{Type versus token embeddings} Tables \ref{tab:results} and \ref{tab:systems} illustrate the gap in performance between type-based embedding models and the token-based ones. Out of the best 10 systems in Subtask 1/Subtask 2, 7/8 systems are based on type embeddings compared to only 2/2 systems that are based on token embeddings (same holds for each language individually). Contrary to the recent success of token embeddings \cite{peters-etal-2018-deep} and to commonly held view that contextual embeddings ``do everything better", they are overwhelmingly outperformed by type embeddings in our task. This is most surprising for Subtask 1, because type embeddings do not distinguish between different senses, while token embeddings do. We suggest several possible reasons for these surprising results. The first is the fact that contextual embedding is a recent technology, and as such lacks proper usage conventions. For example, it is not clear whether a model should create an average token representation based on individual instances (and if so, which layers should be averaged), or if it should use clustering of individual instances instead (and if so, what type of clustering algorithm etc.). A second reason may be related to the fact that contextual models are pretrained and cannot exclusively be trained on the relevant historical resources (in contrast to type embeddings). As such, they carry additional, and possibly irrelevant, information that may mask true diachronic changes. The results may also be related to the specific preprocessing we applied to the corpora: (i) Only restricted context is available to the models as a result of the sentence shuffling. Usually, token-based models take more context into account than just the immediate sentence \cite{martinc-etal-2020-context}. (ii) The corpora were lemmatized, while token-based models usually take the raw sentence as input. In order to make the input more suitable for token-based models, we also provide the raw corpora after the evaluation phase and will publish the annotated uses of the target words with additional context.\footnote{\url{https://www.ims.uni-stuttgart.de/data/sem-eval-ulscd}}

\begin{table}
\tabcolsep=0.11cm
\centering
\begin{tabular}{l|rr|rr}
\toprule
\multirow{2}{*}{\textbf{System}}&\multicolumn{2}{c|}{\textbf{Subtask 1}}&\multicolumn{2}{c}{\textbf{Subtask 2}}\\
 &  Avg. &  Max. &  Avg. &  Max. \\
\midrule
type embeddings  &   \textbf{0.625} &     \textbf{0.687} &         0.329 &          \textbf{0.527} \\
ensemble         &   0.621 &    0.621 &         \textbf{0.442} &          0.442 \\
token embeddings &   0.598 &     0.637 &         0.258 &          0.374 \\
topic model      &   0.526 &     0.526 &         0.100 &          0.100 \\
graph            &   0.521 &     0.521 &         0.234 &          0.234 \\
\bottomrule
\end{tabular}
\caption{Average and maximum performance of best submissions per subtask for different system types. Submissions that corresponded exactly to the baselines or the sample submission were removed.}\label{tab:systems}
\vspace{-2ex}
\end{table}

\begin{figure}
\begin{subfigure}{.5\textwidth}
  \centering
  \includegraphics[width=.95\linewidth]{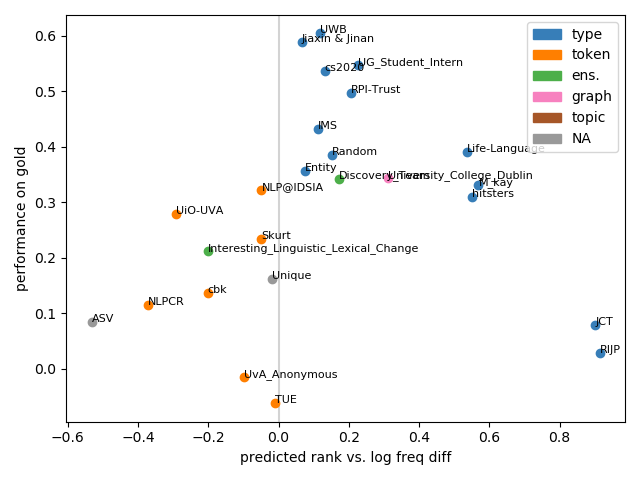}  
  \label{fig:sub-second}
\end{subfigure}
\begin{subfigure}{.5\textwidth}
  \centering
  \includegraphics[width=.95\linewidth]{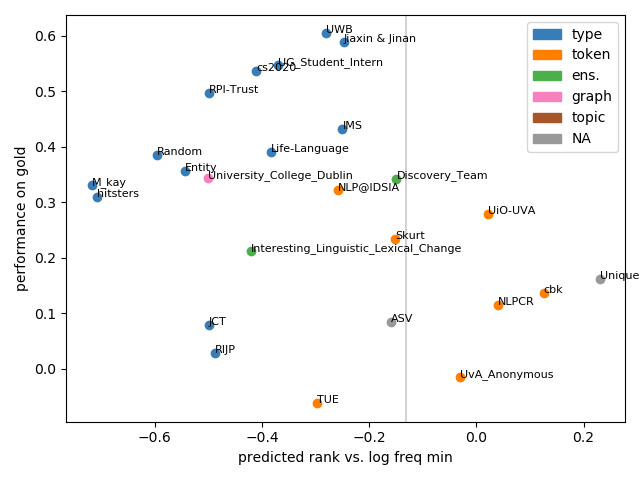}  
  \label{fig:mean_error_latin}
\end{subfigure}
\caption{Influence of frequency on model predictions in Subtask 2, Swedish. X-axis: correlations with FRQ\textsubscript{d} (left) and FRQ\textsubscript{m} (right), Y-axis: performance on Subtask 2. Gray line gives frequency correlation in gold data.}
\label{fig:corrcorr}
\vspace{-8ex} 
\end{figure}

\paragraph{The influence of frequency} In prior work, the predictions of many systems have been shown to be inherently biased towards word frequency, either as a consequence of an increasing sampling error with lower frequency \cite{dubossarsky2017} or by directly relying on frequency-related variables \cite{schlechtweg-EtAl:2017:CoNLL,Schlechtwegetal19}. We have controlled for frequency when selecting target words (recall Table \ref{tab:targetstats}) in order to test model performance when frequency is not an indicating factor. Despite the controlled test sets we observe strong frequency biases for the individual models as illustrated for Swedish in Figure \ref{fig:corrcorr}.\footnote{Find the full set of analysis plots at \url{https://www.ims.uni-stuttgart.de/data/sem-eval-ulscd-post}.} 
Models rather correlate negatively with the minimum frequency of target words between corpora (FRQ\textsubscript{m}), and positively with the change in their frequency across corpora (FRQ\textsubscript{d}). This means that models predict higher change for low-frequency words and higher change for words with strong changes in frequency. Despite their superior performance, type embeddings are more strongly influenced by frequency than token embeddings, probably because the latter are not trained on the test corpora limiting the influence of frequency. Similar tendencies can be seen for the other languages. For a range of models correlations reach values $> 0.8$.

\paragraph{The influence of polysemy} We did not control the test sets for polysemy. As shown in Table \ref{tab:targetstats}, the change scores for both subtasks are moderately to highly correlated with polysemy (PLY\textsubscript{m}). Hence, it is expected that model predictions would be positively correlated with polysemy. However, these are in almost all cases lower than for the change scores and in some cases even negative (Latin and partly English). We conclude that model predictions are only moderately biased towards polysemy on our data.

\begin{figure}
\begin{subfigure}{.5\textwidth}
  \centering
  \includegraphics[width=.95\linewidth]{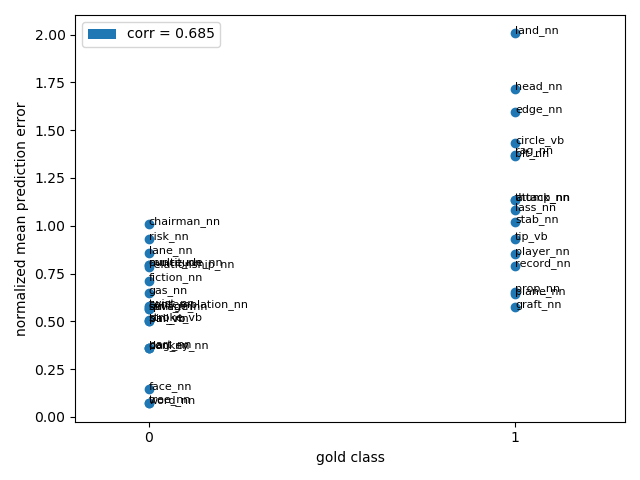} 
  \label{fig:MPE-st1-EN}
\end{subfigure}
\begin{subfigure}{.5\textwidth}
  \centering
  \includegraphics[width=.95\linewidth]{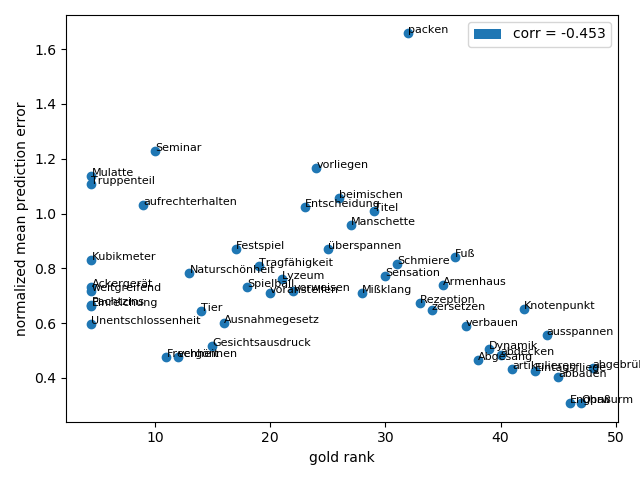}  
  \label{fig:MPE-st2-GE}
\end{subfigure}
\caption{Normalized prediction errors for Subtask~1, English (left) and Subtask~2, German (right).}
\label{fig:corrcorr2}
\vspace{-8ex} 
\end{figure}

\paragraph{Prediction difficulty of words} In order to quantify how difficult a target word is to predict we compute the mean
error of all participants' predictions.\footnote{Because Subtask 2 is a ranking task, we divide the mean error by the expected error: since words in the middle have a lower expected error than words in the top or bottom.} In Subtask 1, we find that words with higher rank tend to have higher error, in particular for English, see Figure~\ref{fig:corrcorr2} (left) where words with the gold class 1 have almost twice as high average error than words with gold class 0, and Latin. This is likely due to the tendency for systems to provide zero-predictions following the published baselines. For Subtask 2 (right), we find that the opposite holds; stable words are harder to predict for all languages but Swedish, where instead, it seems that the words in the middle of the rank are the hardest to classify. For English, the top three hardest to predict words are for Subtask 1 vs. Subtask 2 are \textit{land}, \textit{head}, \textit{edge} vs. \textit{word}, \textit{head}, \textit{multitude}. For German, they are \textit{packen}, \textit{überspannen}, \textit{abgebrüht} vs. \textit{packen}, \textit{Seminar}, \textit{vorliegen}. For Latin, they are \textit{cohors}, \textit{credo}, \textit{virtus} vs. \textit{virtus}, \textit{fidelis}, \textit{itero}. For Swedish, they are \textit{kemisk}, \textit{central}, \textit{bearbeta} vs. \textit{central}, \textit{färg}, \textit{blockera}. We could not identify a general pattern with regards to these words' frequency or polysemy properties.

\section{Conclusion}

We presented the results of the first shared task on Unsupervised Lexical Semantic Change Detection. A wide range of systems were evaluated on two subtasks in four languages relying on a thoroughly annotated data set based on $\sim$100,000 human judgments. The task setup (unsupervised, no genuine development data, different corpora from different languages with very different sizes, varying class distributions) provided an opportunity to test models in heterogeneous learning scenarios, that was very challenging. Hence, both subtasks remain far from solved. However, several teams reach high performances on both subtasks. Surprisingly, type embeddings outperformed token embeddings on both subtasks. We suspect that the potential of token embeddings has not yet fully unfolded, as no canonical application concept is available and preprocessing was not optimal for token embeddings. We found that type embeddings are strongly influenced by frequency. Hence, one important challenge for future type-based models will be to avoid the frequency bias stemming from the corpus on which they are trained. An important challenge for token-based models will be to understand the reasons for their current low performance and to develop robust ways for their application. We found that change scores in our test sets strongly correlate with polysemy, despite model predictions not showing such strong influence. 
We believe that this should be pursued in the future by controlling test sets for polysemy.

We hope that SemEval-2020 Task~1 makes a lasting contribution to the field of Unsupervised Lexical Semantic Change Detection by providing researchers with a standard evaluation framework and high-quality data sets. Despite the limited size of the test sets, many previously reached conclusions can now be tested more thoroughly and future models can be compared on a shared benchmark. The current test set can also be used to test models that have been trained on the full data available for the participating corpora. Data from additional time periods can be utilized by models that need finer granularity for detection, while testing on the two time periods available in the current test sets.

\section*{Acknowledgments}
The authors would like to thank Dr. Diana McCarthy for her valuable input to the genesis of this task. DS was supported by the Konrad Adenauer Foundation and the CRETA center funded by the German Ministry for Education and Research (BMBF) during the conduct of this study. This task has been funded in part by the project \textit{Towards Computational Lexical Semantic Change Detection} supported by the Swedish Research Council (2019–2022; dnr 2018-01184), and \emph{Nationella språkbanken} (the Swedish National Language Bank) -- jointly funded  by  (2018--2024; dnr 2017-00626) and its 10 partner institutions, to NT. The list of potential change words in Swedish was provided by the research group at the Department of Swedish, University of Gothenburg that works with the Contemporary Dictionary of the Swedish Academy. This work was supported by The Alan Turing Institute under the EPSRC grant EP/N510129/1, to BMcG. Additional thanks go to the annotators of our datasets, and an anonymous donor. 

\bibliographystyle{coling}
\bibliography{biblio}

\newpage
\appendix

\section{Annotation Details}
\label{sec:example}

\subsection{Edge sampling} 
\label{sec:edge2}

Retrieving the full usage graph is not feasible even for a small set of $n$ uses as this implies annotating $n*(n-1)/2$ edges. Hence, the main challenge with our annotation approach was to reduce the number of edges to annotate as few as possible, while keeping the necessary information needed to infer a meaningful clustering on the graph. We did this by annotating the data in several rounds. After each round the usage graph of a target word was updated with the new annotations and a new clustering was obtained.\footnote{If an edge was annotated by several annotators we took the median as an edge weight.} Based on this clustering we sampled the edges for the next round applying simple heuristics similar to \newcite{Biemann-2013-system}. We spread the annotation load randomly over annotators making sure that roughly half of the use pairs is annotated by more than one annotator. 

In the first round we aimed to obtain a small but good reference set of uses which would serve to compare the rest of uses in the second round. Hence, we sampled 10\% of the uses from $U$ and 30\% of the edges from this sample by \textbf{exploration}, i.e., by a random walk through the sample graph guaranteeing that all nodes are connected by some path. Hence, the first clustering was obtained on a small but richly connected subgraph guaranteeing that we did not infer a larger number of clusters than present in the data in the first round, which would lead to a strong increase in annotation instances in the subsequent rounds. In all subsequent rounds we combined a \textbf{combination} step with an exploration step. A multi-cluster is a cluster with $\geq 2$ uses. The combination step combined each single use $u_1$ which is not yet member of a multi-cluster with a random use $u_2$ from each of the multi-clusters to which $u_1$ had not yet been compared. The exploration step consisted of a random walk on 30\% of the edges from the non-assignable uses, i.e., uses which had already been compared to each of the multi-clusters but were not assigned to any of these by the clustering algorithm. This procedure slowly populated the graph while minimizing the annotation of redundant information. The procedure stopped when each cluster had been compared to each other cluster. We validated the procedure in a simulation study (see below).

\begin{inparaenum}[(i)]
We combined the above procedure with further heuristics added after round 1: \item we sampled a low number of randomly chosen edges and edges between already confirmed multi-clusters for further annotation to corroborate the inferred structure; \item we detected relevant disagreements between annotators, i.e., judgments with a difference of $\geq 2$ on the scale or edges with a median $\approx 2.5$, and redistributed the corresponding edges to another annotator to resolve the disagreements; and \item we detected clustering conflicts, i.e., positive edges between clusters and negative edges within clusters (see below) and sampled a new edge for each node connected by a conflicting edge. This added more information in regions of the graph where finding a good clustering was hard. Furthermore, after each round, we removed nodes from the graph whose 0-judgments (undecidable) made up more than half of their total judgments. We stopped the annotation after four rounds.
\end{inparaenum}

\subsection{Example} 

\begin{figure}[]
    \begin{subfigure}{0.33\textwidth}
\frame{        \includegraphics[width=\linewidth]{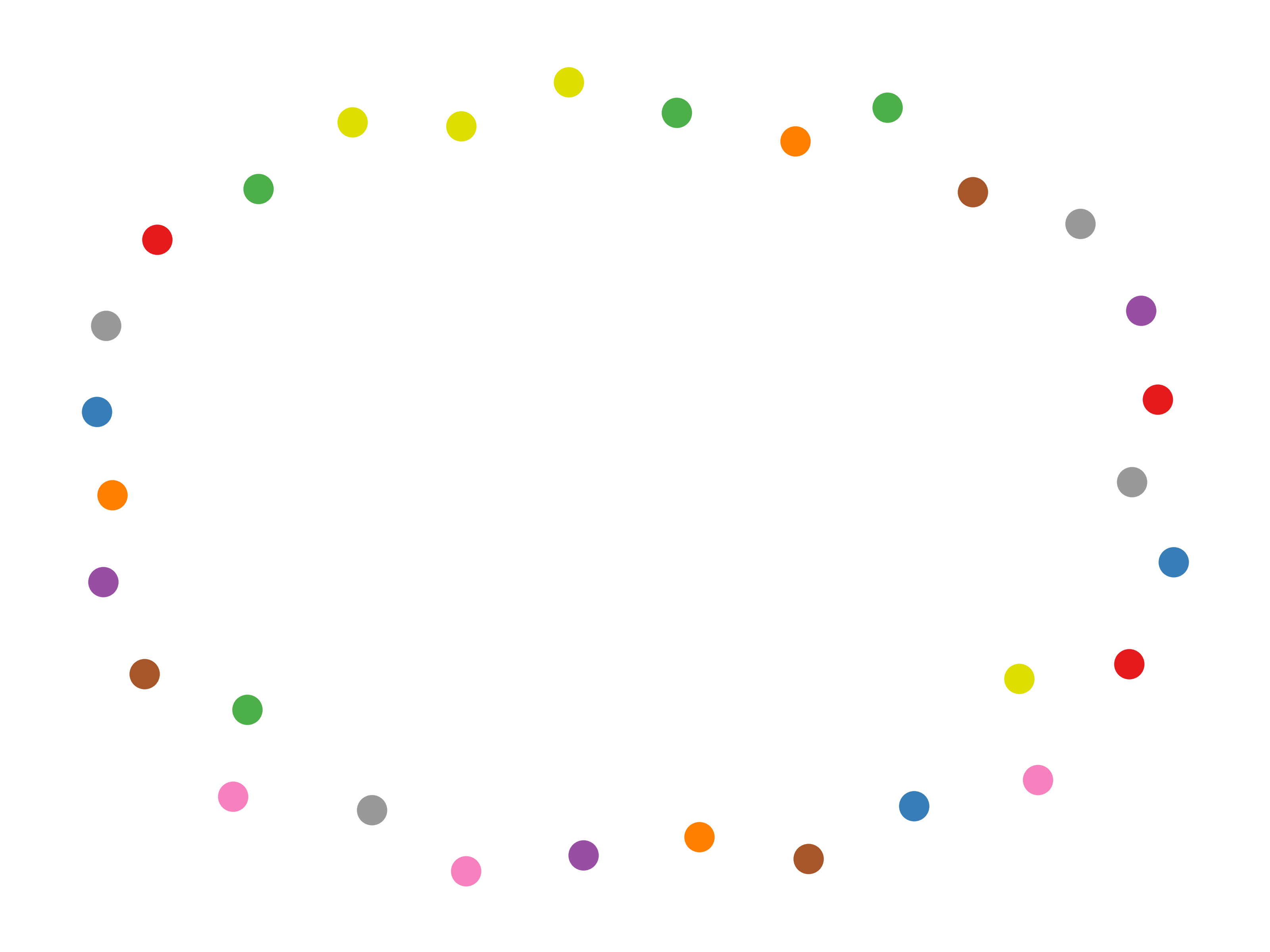}}
        \caption*{round 0}
    \end{subfigure}
    \begin{subfigure}{0.33\textwidth}
\frame {        \includegraphics[width=\linewidth]{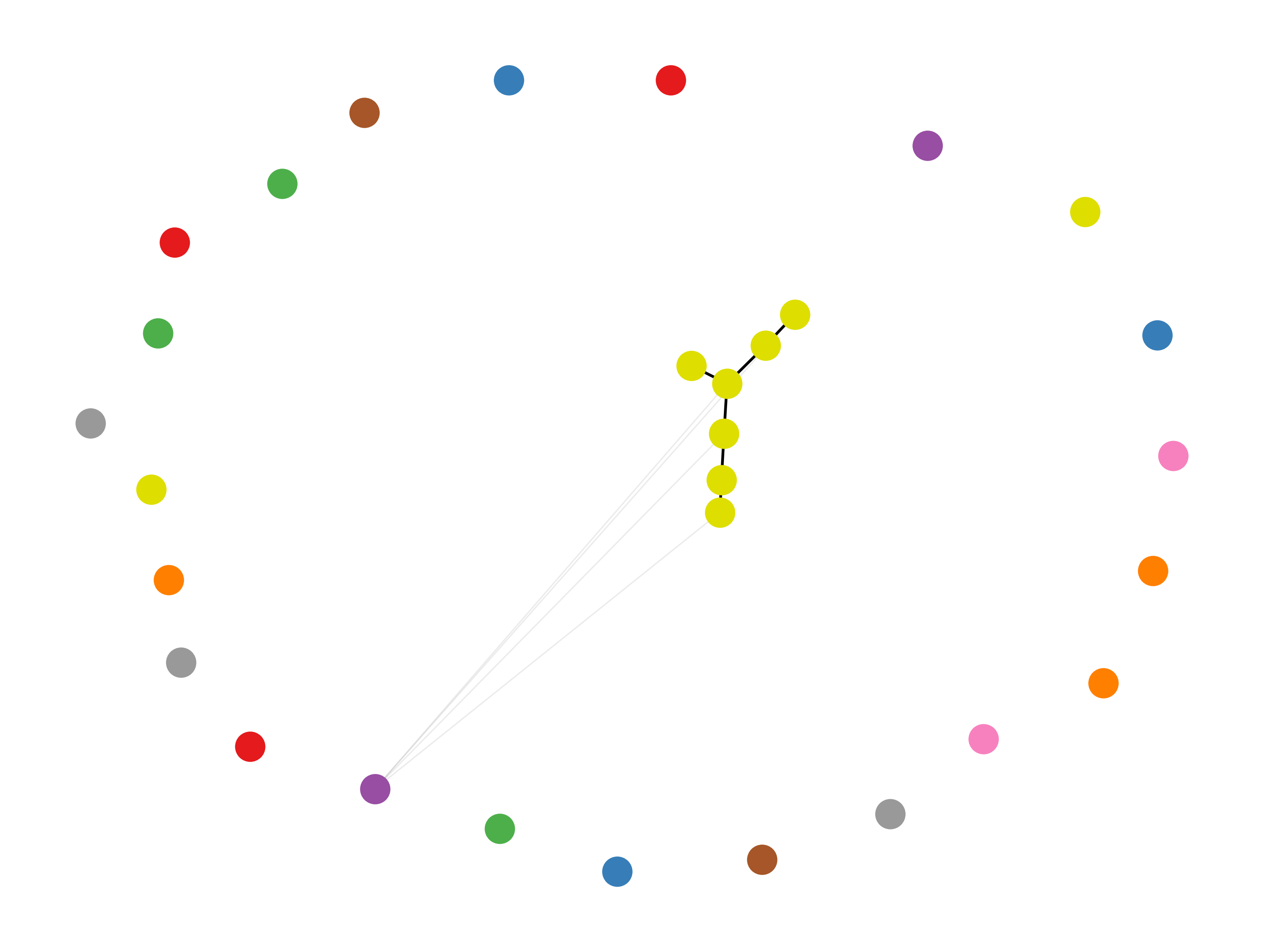}}
        \caption*{round 1}
    \end{subfigure}
    \begin{subfigure}{0.33\textwidth}
\frame {        \includegraphics[width=\linewidth]{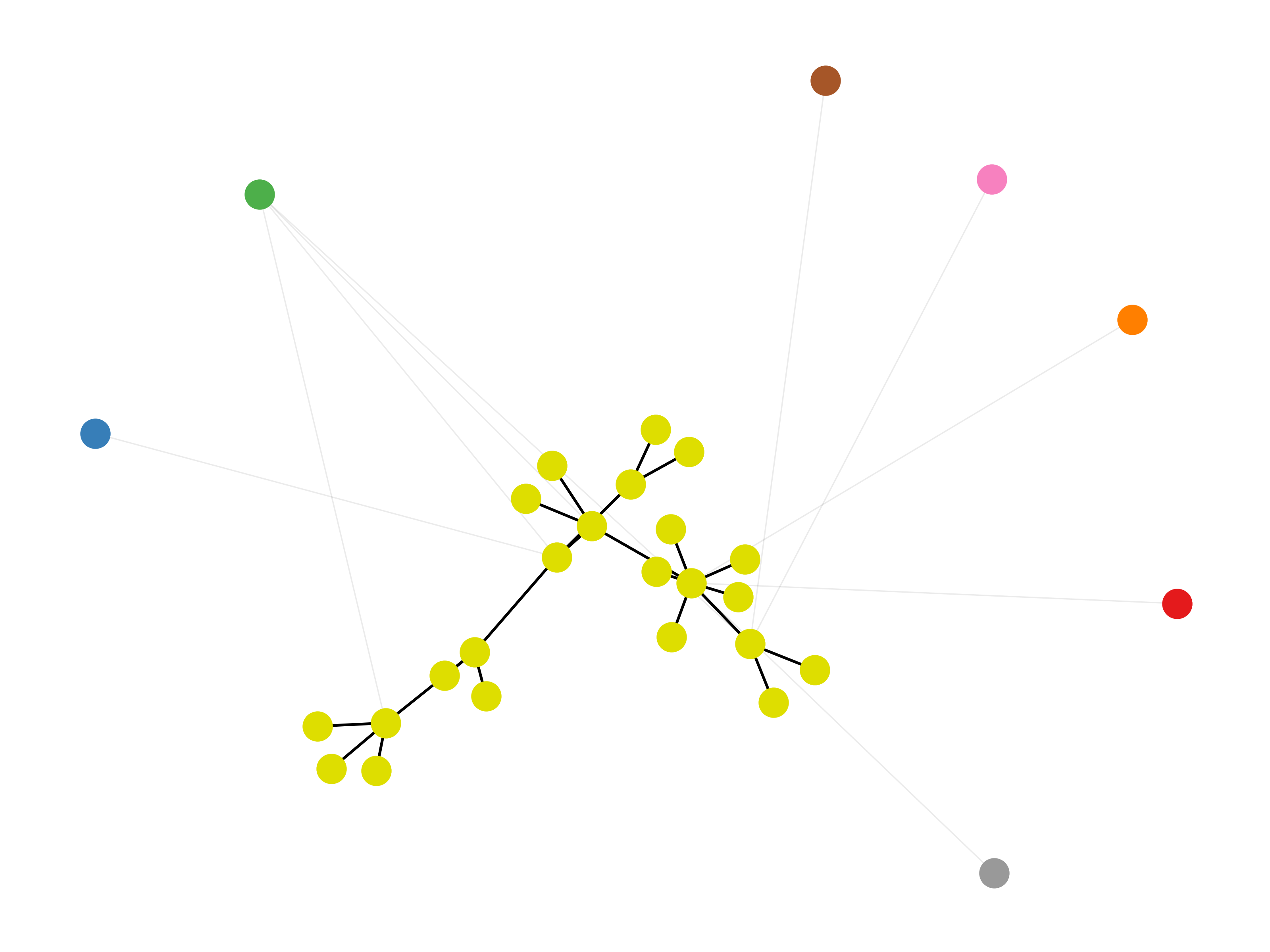}}
        \caption*{round 2}
    \end{subfigure}
    \begin{subfigure}{0.33\textwidth}
\frame {        \includegraphics[width=\linewidth]{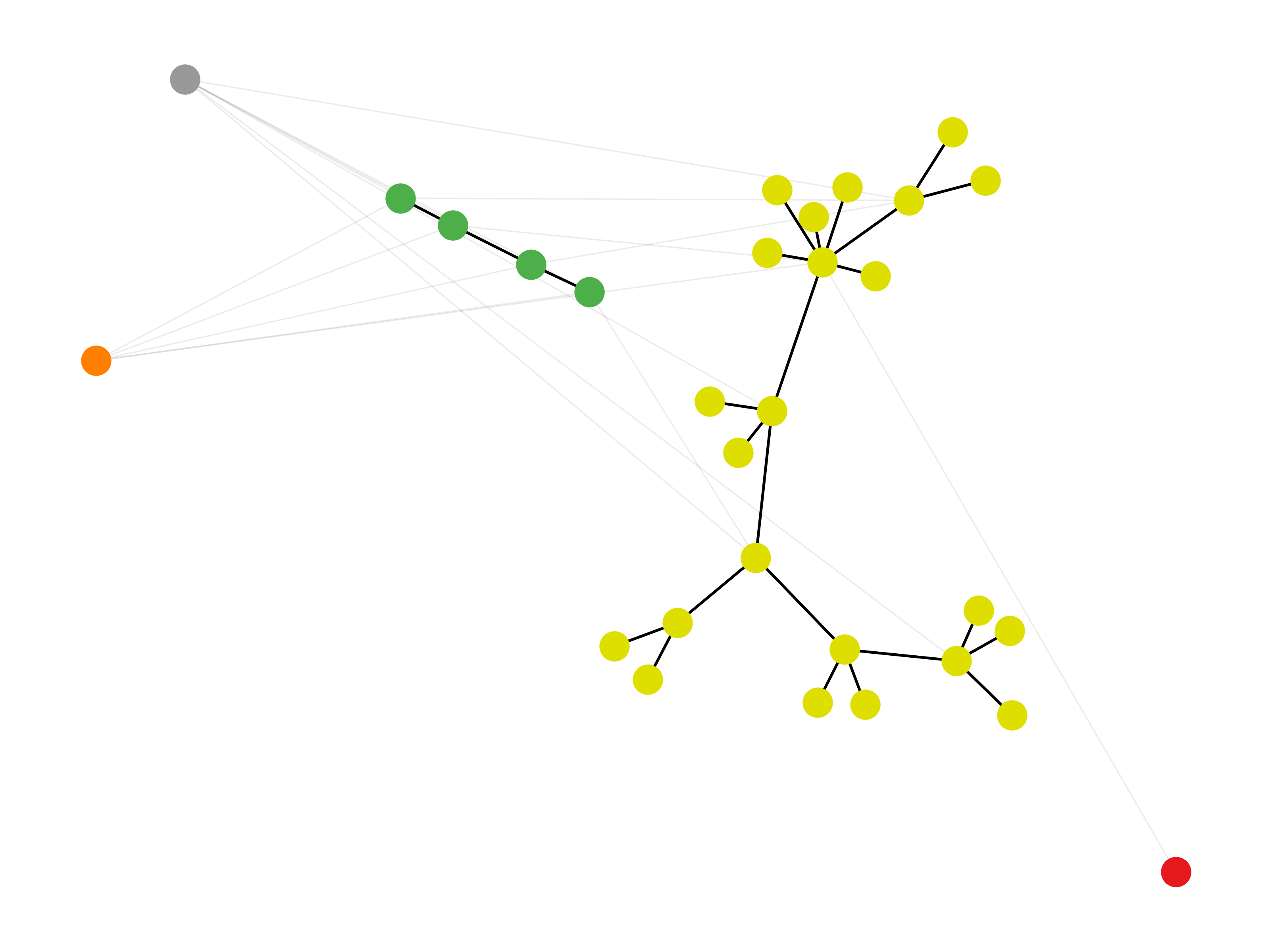}}
        \caption*{round 3}
    \end{subfigure}
    \begin{subfigure}{0.33\textwidth}
\frame {        \includegraphics[width=\linewidth]{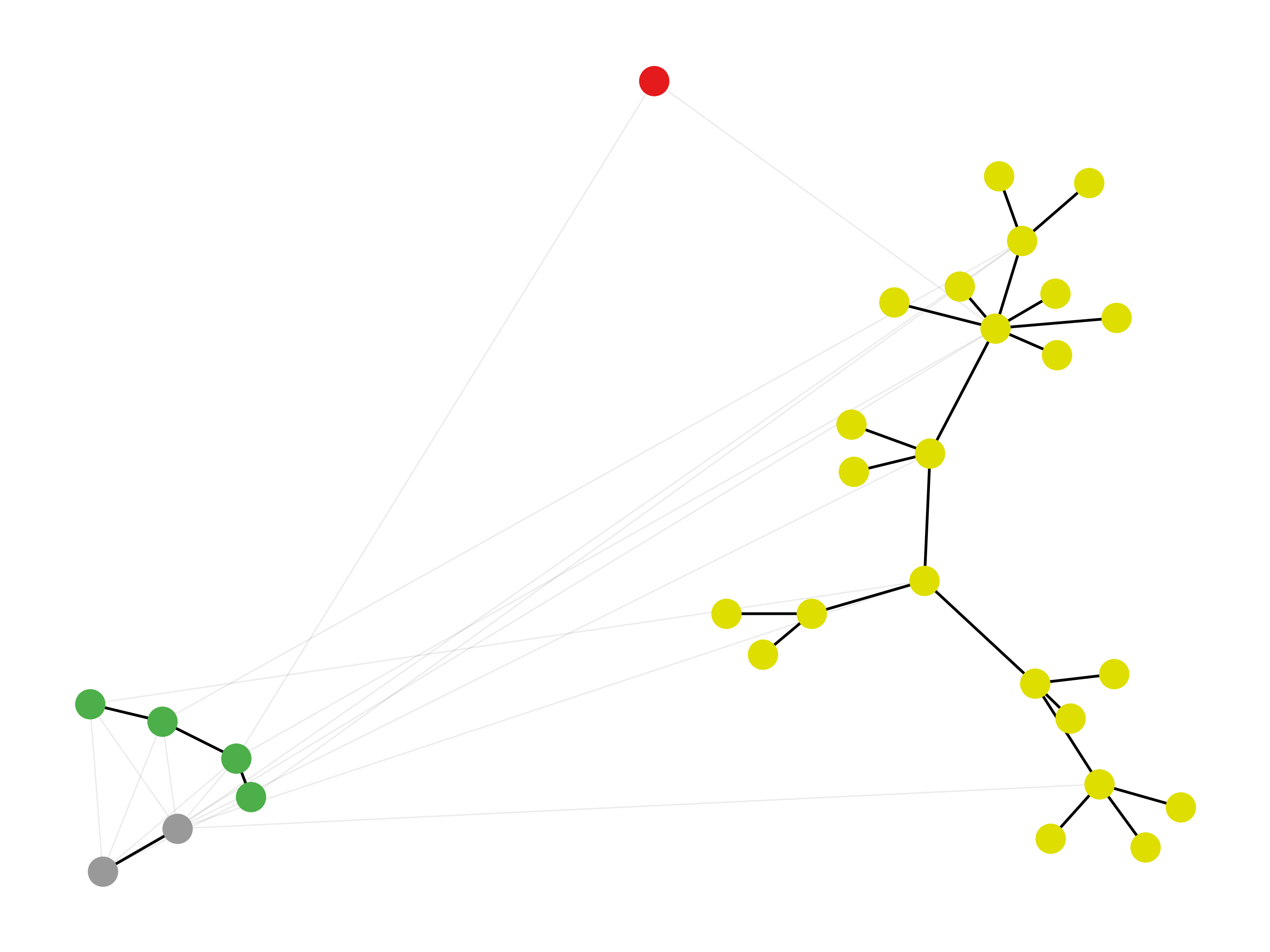}}
        \caption*{round 4}
    \end{subfigure}
    \begin{subfigure}{0.33\textwidth}
\frame {        \includegraphics[width=\linewidth]{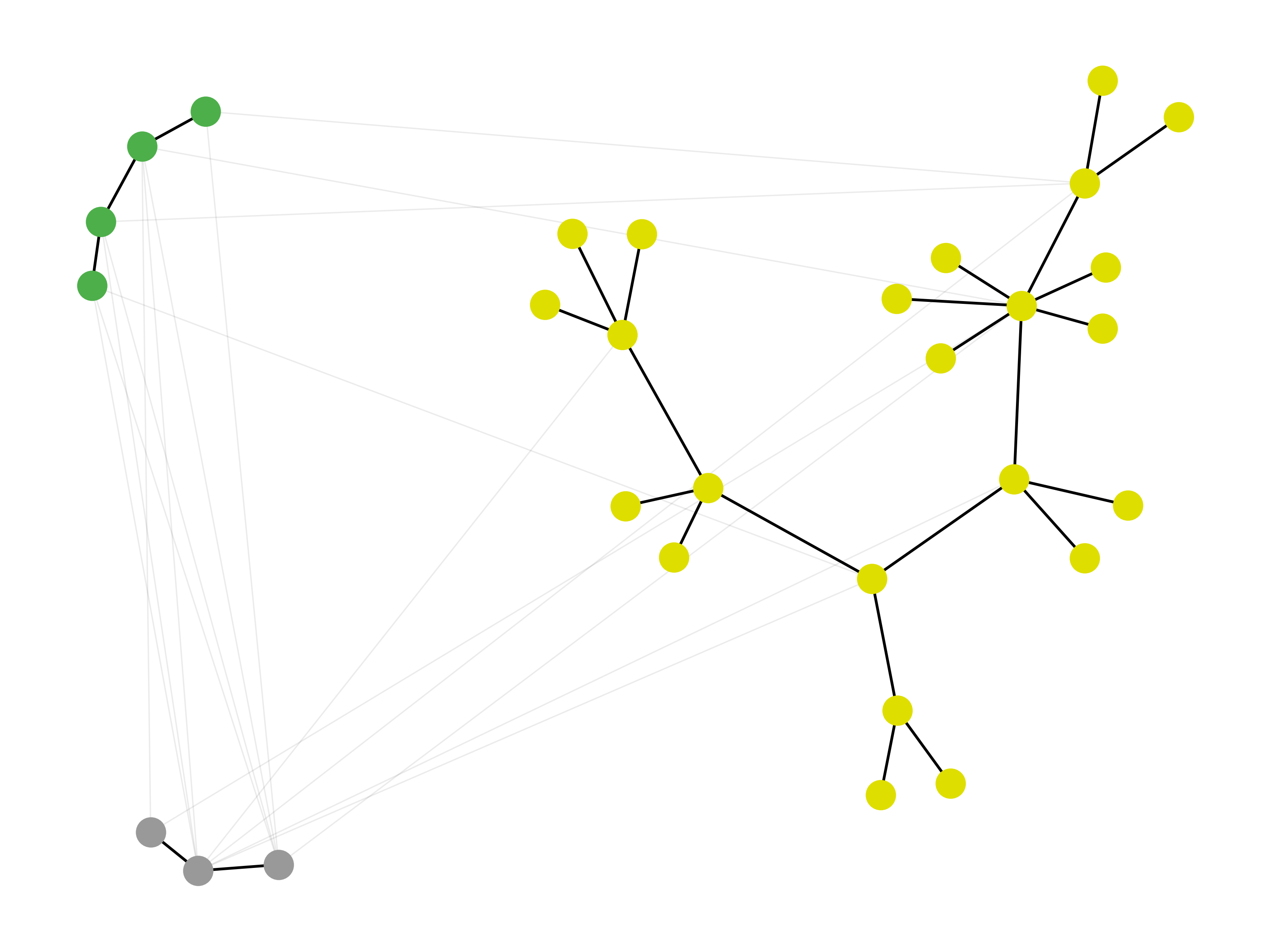}}
        \caption*{round 5}
    \end{subfigure}
    \caption{Simulated example of annotation pipeline.}\label{fig:example}
\end{figure}

Find an example of our annotation pipeline in Figure \ref{fig:example}. As the annotation proceeds through the rounds the graph becomes more populated and the true cluster structure is found. In round 1 one multi-cluster is found. Hence, all remaining uses are compared with this cluster in round 2 by the combination step. In rounds 3 and 4 the exploration step discovers more clusters not found in the rounds before.

\subsection{Simulation} 

We validated the annotation procedure and the clustering algorithm described in Section \ref{sec:annotation} in a simulation study by simulating 40 ground truth usage graphs with zipfian sense frequency distributions covering roughly the frequency range of the majority our target words (50--1000). We introduced change to half of the target words by setting some of its senses' frequencies to 0 in either of $D_1$, $D_2$. We then sampled from these graphs in several rounds as described above, simulated an annotation in each round with a normally distributed error added to judgments and compared the resulting clustering to the clustering of the true graph. The true clustering could be recovered with high accuracy (average of $>.96$ adjusted mean rand index). We also used the simulation to predict the feasibility of the study and to tune parameters of the annotation such as sample sizes for nodes and edges. With the finally chosen parameters described in Section \ref{sec:edge} the algorithm converged on average after 5 rounds and $\approx 8000$ judgments per annotator. This was within the bounds of our time limits and financial budget.

We also tested the clustering algorithm against several standard techniques \cite{Biemann2006,Blondel2008} and varied the optimization algorithm for $L$. None of these variations performed compatible with our approach.

\section{Systems description}
\label{A:systems}

\paragraph{cbk \cite{beck-2020-diasense}} The team obtains contextual embeddings using BERT \cite{devlin-etal-2019-bert}, and extracts for every target word usage a word embedding using bert-as-service \cite{xiao2018bertservice}. The team uses the difference of mean value of all cosine distances \cite{salton1986introduction} of a target word between two corpora to detect change.

\paragraph{cs2020\footnote{The team is named \enquote{cs2020} and \enquote{cs2021} on the competition CodaLab. The combined number of submissions made by the two teams did not exceed the limit of 10.} \cite{arefyev-zhikov-2020-cmce}}
The team submits systems of two types: SGNS with an Orthogonal Procrustes alignment and cosine distance as a change measure, and a variation of a word-sense induction method by \newcite{amrami2018word}. 
For the latter, the team replaces BERT by a fine-tuned version of XLM-R \cite{conneau2019unsupervised} and for every target word generates lexical substitutes following \newcite{amrami2019towards}, the vectors of the most probable of which are then clustered using agglomerative clustering, with cosine distance. 

\paragraph{Discovery Team \cite{martinc-etal-2020-context}} The team uses two types of word representations: average embeddings from SGNS \cite{Mikolov13a,Mikolov13b} with an Orthogonal Procrustes alignment and contextual embeddings using language-specific BERT \cite{devlin-etal-2019-bert}. For SGNS+OP the team compares vectors using cosine \cite{salton1986introduction}, while contextual embeddings see two different strategies: averaging of target-word embeddings, and clustering using k-means and affinity propagation \cite{frey2007clustering}. Clusters are then compared across time using Jensen-Shannon divergence. The team also submits an ensemble model that uses all four strategies (SGNS+OP, averaging, k-means, AP).

\paragraph{DCC \cite{zamora-reina-etal-2020-dcc-uchile}}
This team's system was designed for Subtask 1 and is based on Temporal Referencing \cite{Dubossarskyetal19}.
They concatenated the corpora of different periods and treated the occurrences of target words in different periods as two independent tokens. Afterwards, they trained word embeddings on the joint corpus and compare the referenced vectors of each target word using cosine similarity.
They used the Gensim package for training word2vec embeddings (Continuous Bag of Words with Negative Sampling). The number of negative samples was set to 5.

\paragraph{Entity \cite{jain-2020-gloveinit}} The team obtains GloVe average embeddings \cite{pennington2014glove}, then applies vector initialization (VI) alignment \cite{Kim14} and measures change with cosine distance \cite{salton1986introduction}.

\paragraph{IMS\footnote{The team is named \enquote{in vain} on the competition CodaLab.} \cite{kaiser-etal-2020-IMS}} The team obtains average embeddings from Skip-Gram with Negative Sampling \cite[SGNS]{Mikolov13a,Mikolov13b}, then applies vector initialization (VI) alignment \cite{Kim14} and measures change with cosine distance \cite{salton1986introduction}. (SGNS+VI+CD) VI is an alignment strategy where the vector space learning model for $C_2$ is initialized with the vectors from $C_1$. They take a cutoff approach to Subtask 1. 

\paragraph{Jiaxin \& Jinan\footnote{The team is named \enquote{LYNX} on the competition CodaLab.} \cite{zhou-etal-2020-temporalteller}} The team uses two static word embedding models, SGNS and PPMI, in addition to averaged BERT representation, with different methods for alignment (Orthogonal Procrustes, Temporal Referencing, and Column Intersection). They compute the cosine distance between these representations and use it standardly for Subtask-2. For Subtask-1 they propose the Gamma Quantile Threshold as a novel approach too choose the change cutoff based on cosine-distance distributions. The team also provides a large scale comparison of two hyper-parameters, embedding dimensionality and context size window.

\paragraph{JCT \cite{amar-liebeskind-2020-jct}}
The team systematically combined existing models for lexical semantic change detection,\footnote{They considered various  word representations (raw count vectors, positive Pointwise Mutual Information, Singular Value Decomposition, Random Indexing, and Skip-Gram with Negative Sampling), alignment methods (Column Intersection, Shared Random Vectors, Orthogonal Procrustes, Vector Initialization), and similarity and dispersion measures (cosine distance, local neighbourhood distance, frequency difference, type difference, and entropy difference).
}
and analyzed their score distribution. After defining a general threshold in an unsupervised way, they adjusted it to each of the models. They measured the models' decision certainty and used it to filter the best models; they then applied the majority rule to select the binary class for Subtask 1 and used a weighted average of the scores of the best models to obtain the ranking for Subtask 2. 

\paragraph{Life-Language \cite{asgari-etal-2020-emblexchange}} The team trains static embedding (fasttext) on each corpus separately. Target words are represented as probability distributions (after softmax) by their distance relative to fixed pivot words that are considered to be highly stable (according to some frequency heuristics). Semantic change is defined as the KL-divergence between two distributions of the target word. An empirical semantic change distribution is computed by randomly choosing pivot words. This enables to choose a threshold for the change score (Subtask 1). For Subtask 2, the change score is straightforwardly used for ranking.

\paragraph{NLP@IDSIA\footnote{The team is named \enquote{Vani Kanjirangat} on the competition CodaLab.} \cite{kanjirangat-etal-2020-sst}} The team obtains contextual embeddings using BERT \cite{devlin-etal-2019-bert}.
The team uses Euclidean distance between vectors of word uses to create clusters using k-means, and the silhouette method to define the number of clusters. They cluster word vectors using two strategies: joined corpora, and separately (within-corpus clusters).

\paragraph{NLPCR \cite{rother-etal-2020-cmce}} The team uses multilingual contextualized word embeddings \cite{devlin-etal-2019-bert} to represent a word's meaning. They then reduce the embedding dimensionality with either autoencoder or UMAP, and cluster the resulting representation with either GMM or HDBSCAN. For Subtask 1 they use the task's specification directly on the cluster assignments, and for Subtask 2 they use the Jensen-Shannon Divergence for ranking.

\paragraph{Random \cite{cassotti-etal-2020-gmctsc}}
This team focused on the problem of identifying when a target word has gained or lost senses. They train dynamic word embeddings using methods based on both explicit alignment such as Dynamic Word2Vec \cite{yaoetal2018}, and implicit alignment, like Temporal Random Indexing
\cite{basileetal2015} and Temporal Referencing \cite{Dubossarskyetal19}. They also use different similarity measures to determine the extent of a word semantic change and compare the cosine similarity with Pearson Correlation and the neighborhood similarity \cite{shoemarketal2019}. They introduce a new method to classify changing vs. stable words by clustering the target similarity distributions via Gaussian Mixture Models.

\paragraph{RIJP \cite{iwamoto-yukawa-2020-rijp}} The team uses Gaussian embedding \cite{vilnis2014word} to represent words distributions instead of a points as in standard embedding models. Mean vectors are learned with word2vec using \cite{Kim14} method. Covariance matrices are not trained, but encode the words' frequency changes between two time points. Kullback-Leibler (KL) divergence is then applied to measure changes to a word's distribution. 

\paragraph{RPI-trust \cite{gruppi-etal-2020-schme}} The team uses static word embedding aligned using Orthogonal Procrustes. They extract three features from these representations, cosine-distance between words in two time points, change to their nearest-neighbours and frequency change, which they use in an ensemble model. They adopt an anomaly detection approach to find the threshold for change words (Subtask1), and directly computing the rank on the ensemble score (Subtask 2).

\paragraph{Skurt \cite{gyllensten-etal-2020-sensecluster}} The team uses pretrained cross-lingual contextualized embedding model, XLM-R \cite{conneau2019unsupervised}, which enables them to use the same model for all languages. For each target word they generated contextual representations (token representations) from the two corpora, and cluster them using K-Means++ with a fixed number of clusters (8). They use these cluster assignments as a proxy for the words' senses, and compare their between the two corpora. For Subtask 1, they directly implement the criterion provided in the task reference for LSC, and for Subtask 2 they use the Jensen-Shannon Divergence as a score for the degree of change in these senses.

\paragraph{TUE\footnote{The team is named \enquote{Schwaebischschwaetza\_tue} on the competition CodaLab.} \cite{parnysheva-schwarz-2020-tue}} The team uses a cross-lingual pretrained contextual word embedding model \cite{che2018towards} to represent a word's meaning across the two corpora. They then cluster these representations using either K-Means or DBSCAN, and compare the words' cluster assignments between the two time points. These cluster assignments allows that to tackle Subtask 1 directly (using the criterion defined by the organizers), and to compute Jensen-Shannon Divergence for Subtask 2.

\paragraph{UG Student Intern \cite{pomsl-lyapin-2020-circe}} The team submits three types of model: average embeddings from SGNS \cite{Mikolov13a,Mikolov13b} with an Orthogonal Procrustes alignment with vector comparison through Euclidean distance, contextual embeddings using BERT \cite{devlin-etal-2019-bert} and a sentence time classification objective, and finally their ensemble model CIRCE. 

\paragraph{UiO-UVA \cite{kutuzov-giulianelli-2020-uiouva}} The team obtains contextual embeddings of two types: ELMo \cite{peters-etal-2018-deep} and BERT \cite{devlin-etal-2019-bert}.
The team measures change in three different ways: cosine distance \cite{salton1986introduction} on averaged vectors, average pairwise distance between all contextual vectors of a same target, and Jensen-Shannon divergence applied on clusters created with affinity propagation \cite{frey2007clustering}.

\paragraph{University College Dublin \cite{nulty-lillis-2020-ucdnet}} The team represents words as nodes in a weighted undirected graph that represents their word associations in the two time points. Similar to the Temporal-Referencing model \cite{Dubossarskyetal19}, the nodes represent all words in both time points, and only target words have two nodes. The edges' weights are determined by the words' ppmi scores that surpass a threshold. Resistance distance metric is used to evaluate the degree of change of the target words. For Subtask 1 a threshold is manually set to determine the change and stable words, and for Subtask 2 the distance metric is used straightforwardly in the ranking.

\paragraph{UoB\footnote{The team is named \enquote{Eleri Sarsfield} on the competition CodaLab.} \cite{sarsfield-madabushi-2020-uob}}
The team uses a topic model (Hierarchical Dirichlet processes \cite[HDP]{teh2004sharing}) to model senses for each word, and mimics \newcite{cook2014novel} in utilising the novelty score as a similarity measure.

\paragraph{UWB \cite{prazak-etal-2020-uwb}} The team obtains average embeddings from SGNS \cite{Mikolov13a,Mikolov13b}, and aligns models from the two periods using Canonical Correlation Analysis (CCA), and Orthogonal Procrustes (using VecMap \cite{artetxe-etal-2018-robust}).
The team measures change using cosine distance \cite{salton1986introduction}.

\newpage
\section{Results with F1, Precision and Recall}
\label{A:results}

Find the participants' results on Subtask 1 evaluated with F1, Precision and Recall in Table \ref{tab:results2}.

\begin{table}[htp]
\tabcolsep=0.09cm
\centering
\begin{tabular}{l|lll|lll|lll|lll|lll}
\toprule
{}        &  \multicolumn{3}{c}{\textbf{Avg.}} & \multicolumn{3}{c}{\textbf{English}} &             \multicolumn{3}{c}{\textbf{German}} &              \multicolumn{3}{c}{\textbf{Latin}} &               \multicolumn{3}{c}{\textbf{Swedish}} \\
\midrule
\textbf{Team}&P&R&F1&P&R&F1&P&R&F1&P&R&F1&P&R&F1\\\hline
\midrule
Jiaxin \& Jinan            &  .583& .789& .646 &     .6& .562& .58 &  .583& .824& .683 &  .769& .769& .769 &    .381& 1.0& .552 \\
Skurt                     &     .549& .8& .63 &    .5& .625& .556 &  .441& .882& .588 &  .783& .692& .735 &     .471& 1.0& .64 \\
UWB                       &  .618& .672& .629 &  .556& .625& .588 &    .609& .824& .7 &  .889& .615& .727 &    .417& .625& .5 \\
UG\_Student\_Intern         &    .562& .7& .607 &    .5& .562& .529 &  .583& .824& .683 &    .7& .538& .608 &  .467& .875& .609 \\
\textbf{Freq. Bas.}        &  .426& .971& .579 &    .432& 1.0& .603 &  .366& .882& .517 &     .65& 1.0& .788 &     .258& 1.0& .41 \\
IMS                       &  .523& .654& .564 &  .474& .562& .514 &  .542& .765& .634 &    .7& .538& .608 &     .375& .75& .5 \\
Random                    &   .512& .701& .56 &  .452& .875& .596 &  .395& .882& .546 &  .647& .423& .512 &  .556& .625& .588 \\
Life-Language             &  .683& .487& .559 &   .778& .438& .56 &  .667& .588& .625 &   .786& .423& .55 &        .5& .5& .5 \\
NLPCR                     &   .504& .567& .52 &   .667& .75& .706 &  .381& .471& .421 &    .611& .423& .5 &  .357& .625& .454 \\
Discovery\_Team            &   .593& .499& .51 &    .5& .438& .467 &  .556& .588& .572 &      .9& .346& .5 &    .417& .625& .5 \\
UiO-UvA                   &    .5& .496& .493 &    .471& .5& .485 &    .5& .647& .564 &    .6& .462& .522 &    .429& .375& .4 \\
DCC                       &  .625& .451& .492 &  .714& .312& .434 &  .524& .647& .579 &  .818& .346& .486 &     .444& .5& .47 \\
Entity                    &  .622& .495& .478 &  .833& .312& .454 &  .524& .647& .579 &    .778& .269& .4 &    .353& .75& .48 \\
cs2020                    &  .468& .466& .464 &  .538& .438& .483 &   .267& .235& .25 &  .667& .692& .679 &      .4& .5& .444 \\
RIJP                      &  .475& .462& .429 &  .462& .375& .414 &   .37& .588& .454 &  .833& .385& .527 &     .235& .5& .32 \\
NLP@IDSIA                 &   .571& .361& .42 &   .75& .188& .301 &    .462& .353& .4 &  .739& .654& .694 &   .333& .25& .286 \\
RPI-Trust                 &  .789& .243& .365 &    1.0& .188& .316 &    .857& .353& .5 &    .8& .308& .445 &      .5& .125& .2 \\
UCD &   .51& .481& .362 &   .75& .188& .301 &    .4& .824& .539 &    .5& .038& .071 &  .389& .875& .539 \\
JCT                       &  .767& .226& .337 &    1.0& .188& .316 &  .667& .235& .348 &    1.0& .231& .375 &     .4& .25& .308 \\
\textbf{Count Bas.}        &  .682& .238& .318 &    1.0& .062& .117 &    .625& .294& .4 &  .818& .346& .486 &   .286& .25& .267 \\
cbk                       &    .437& .216& - &      .5& .125& .2 &  .471& .471& .471 &    .778& .269& .4 &        .0& .0& - \\
UoB                       &      -& .401& - &        -& .0& - &  .346& .529& .418 &  .714& .577& .638 &     .25& .5& .333 \\
TUE                       &      -& .341& - &        -& .0& - &    .4& .353& .375 &  .676& .885& .767 &    .2& .125& .154 \\
\textbf{Maj. Bas.}         &        -& .0& - &        -& .0& - &        -& .0& - &        -& .0& - &        -& .0& - \\
\bottomrule
\end{tabular}
\caption{Summary of the precision (P), recall (R), and F1 scores on Subtask 1 for the baseline systems and the systems which submitted a system description paper. `Avg.' refers to the average across all languages for each system. The baseline systems and the submitting systems are ordered by decreasing F1 of their best submission calculated on the average over all languages.}\label{tab:results2}
\end{table}

\end{document}